\documentclass[10pt,twocolumn,letterpaper]{article}

\usepackage{iccv}
\usepackage{times}
\usepackage{epsfig}
\usepackage{graphicx}
\usepackage{amssymb}

\usepackage{amsmath}
\usepackage{graphicx} 
\usepackage{bmpsize}
\usepackage{url}
\usepackage{xcolor}
\usepackage{wrapfig}
\usepackage{subfig}
\usepackage{authblk}
\usepackage{amsfonts,amssymb}

\usepackage[pagebackref=true,breaklinks=true,letterpaper=true,colorlinks,bookmarks=false]{hyperref}

\iccvfinalcopy 

\def\iccvPaperID{8054} 

\begin{document}

\title{Hierarchically Compositional Tasks and Deep Convolutional Networks}

\author{Arturo Deza, Qianli Liao, Andrzej Banburski, Tomaso Poggio\\
Center for Brains, Minds and Machines\\
Massachusetts Institute of Technology\\
{\tt\small deza@mit.edu, lql@mit.edu, kappa666@mit.edu, tp@csail.mit.edu}
}

\vspace{-40pt}
\maketitle


\vspace{-10pt}
\begin{abstract}
\vspace{-10pt}
The main success stories of deep learning, starting with ImageNet, depend on deep convolutional networks, which on certain tasks perform significantly better than traditional shallow classifiers, such as support vector machines, and also better than deep fully connected networks; but what is so special about deep convolutional networks? Recent results in approximation theory proved an exponential advantage of deep convolutional networks with or without shared weights  in approximating functions with hierarchical locality in their compositional structure. More recently, the hierarchical structure was proved to be hard to learn from data, suggesting that it is a powerful prior embedded in the architecture of the network. These mathematical results, however, do not say which real-life tasks correspond to input-output functions with hierarchical locality. To evaluate this, we consider a set of visual tasks where we disrupt the local organization of images via ``deterministic scrambling'' to later perform a visual task on these images structurally-altered in the same way for training and testing. For object recognition we find, as expected, that scrambling does not affect the performance of shallow or deep fully connected networks contrary to the out-performance of convolutional networks. Not all tasks involving images are however affected. Texture perception and global color estimation are much less sensitive to deterministic scrambling showing that the underlying functions corresponding to these tasks are not hierarchically local; and also counter-intuitively showing that these tasks are better approximated by networks that are not deep (texture) nor convolutional (color). Altogether, these results shed light into the importance of matching a network architecture with its embedded prior of the task to be learned. \end{abstract}

\vspace{-15pt}
\section{Introduction}
At the heart of the success of many biological + artificial object recognition systems, is the prior of compositionality: a scene is composed of a collection of objects; an object is composed by a collection of parts; and a face -- arguably the ``most discriminable object'' that we learn to categorize in our lifetime~\cite{kanwisher1997fusiform} -- is a collection of sub-parts such as a pair of eyes, ears, nose and a mouth. Even prior to the 2nd wave of deep learning with the advent of AlexNet's~\cite{krizhevsky2012imagenet} success on ImageNet~\cite{russakovsky2015imagenet}, there have been a series of models that have depended on mixing features in different ways such as using Histogram of Oriented Gradients (HoG)~\cite{dalal2005histograms},  Deformable Parts Model (DPM)~\cite{felzenszwalb2008discriminatively} or Bag-of-Words (BoW)~\cite{csurka2004visual} with a SVM classifier. Why is it then that deep convolutional networks (such as AlexNet) performed so well on object classification tasks compared to their predecessor models, and what could such models do that previously hand-engineered models could not?  Are dense connections worse than sparse ones? Is deeper better than shallow? When do deep convolutional networks fail? The empirical answer to several of these questions is known~\cite{lecun2015deep,goodfellow2016deep}, but theoretical explanations are lacking as they are lacking about issues such as convergence, generalization,  and  adversarial attacks~\cite{zhang2016understanding,poggio2020theoretical,goodfellow2014explaining,geirhos2020shortcut,feather2019metamers}.




Indeed, it seems that the \textit{`unspoken truth'} in the machine learning and computer vision community when talking about these classical results regarding neural network architectures and tasks is that \textit{``deeper and convolutional is better''} -- as they will exploit the compositional and localized structure of images~\cite{henaff2014local}. More specifically, the questions of why and under which conditions however remains:

\vspace{5pt}

\fbox{\begin{minipage}{21em}
 Why and for which tasks do convolutional networks  perform better than fully-connected networks?
\end{minipage}}

\vspace{5pt}

A possible explanation from the point of view of approximation theory was proposed by Poggio~\textit{et~al.}~\cite{poggio2020theoretical}) in theorems that we summarize as follows:

\begin{figure*}[t!]
\centering
\includegraphics[width=1.0\linewidth,clip]{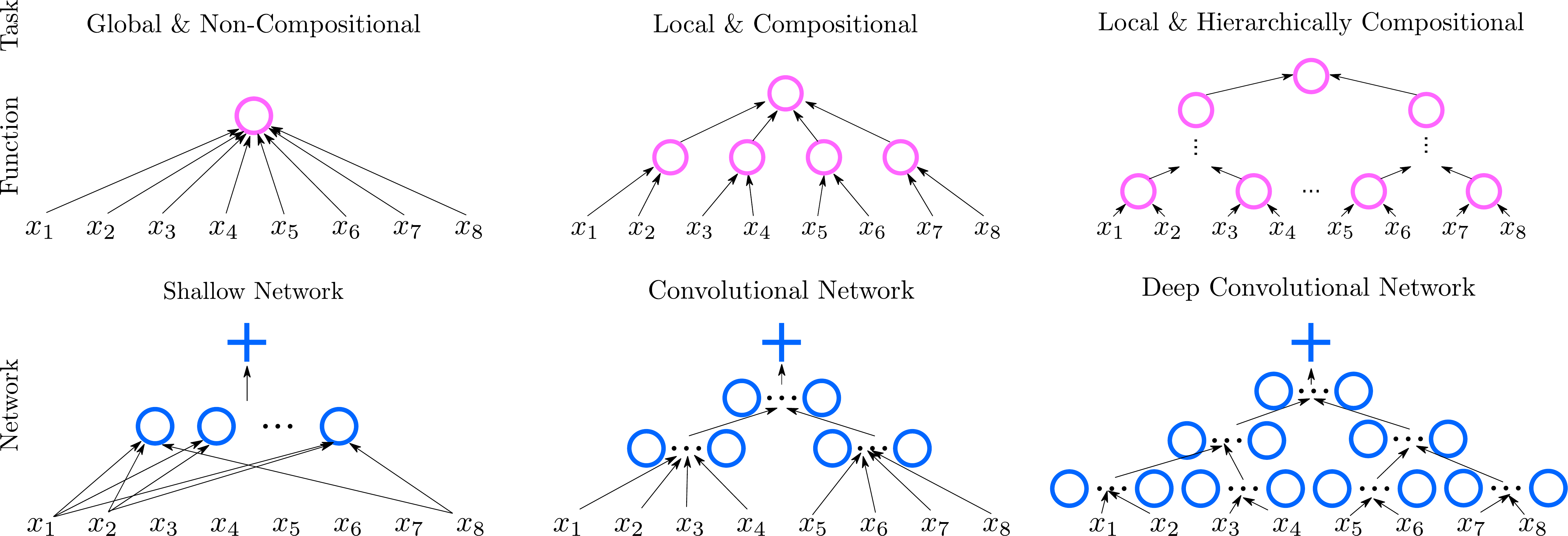}
\caption{A cartoon depicting the different types of tasks (top) in pure mathematical form (center) and their respective `optimal' approximation network (bottom) given a finite input vector: $[x_1, ..., x_8]$. Left: We have a global and non-compositional function -- as it has \textit{no hierarchy nor locality}. Middle: a first order local and compositional function as it relies on \textit{one level} of non-linear local function combination before being sent to a decision function $(+)$. Right: A higher order hierarchically compositional function that consists of a \textit{cascade} of combinations of non-linear local functions. This last group of functions are generally well approximated by deep conv. networks~\cite{Poggio2017}; here we try to understand under for what visual tasks and why}
\label{fig:Network_Architectures}
\end{figure*}


\begin{enumerate}
\itemsep0em 
\item {\it Both shallow and deep networks are universal}, that
  is they can approximate arbitrarily well any continuous function of
  $d$ variables on a compact domain but both suffer from the {\it curse of dimensionality} with a number of parameters of order $O(\epsilon^d)$, where $\epsilon$ is the approximation error.
  
\item For the class of functions of $d$ variables on a
  compact domain that are {\it hierarchical compositions of local (that is with bounded dimensionality) constituent 
    functions}, approximation by
  deep networks with the same architecture  can be achieved with a number of parameters which is linear in $d$ instead than exponential.
  \item The key aspect of convolutional networks that can give them an exponential advantage is not weight sharing
but locality at each level of the hierarchy. Weight sharing helps but not in an exponential way.
\end{enumerate}


The result (1) for shallow networks
  is classical (See Cybenko~\cite{cybenko1989approximation}), and the extension of this result is easy to prove as shallow networks are a subset of deep networks. An example of a function which is a hierarchical compositions of constituent 
    functions is
  $f(x_1, \cdots, x_8) = \phi_3(\phi_{21}(\phi_{11} (x_1, x_2), \phi_{12}(x_3,
  x_4)), \allowbreak \phi_{22}(\phi_{13}(x_5, x_6), \phi_{14}(x_7, x_8))) $. Here the constiuent functions have dimensions $2$. The equivalent quantity in a convolutional network is the dimensionality of the convolution kernel (for images it is typically $3 \times 3$). 
  
  All these results are about the representational power of deep networks and especially hierarchical networks. As such they do not say what \textit{can be} or \textit{cannot be} learned from data. However, the exponential separation between deep convolutional networks and shallow networks for the class of hierarchically local functions suggests that such architectures may be difficult to learn from a reasonable amount of data. Notice that the success of gradient descent techniques in learning overparametrized networks does not imply at all that it is easy to learn convolutional networks from fully connected networks. In fact the opposite is true. Recently, Malach~\&~Shalev-Shwartz~\cite{malach2021computational} demonstrated  a class of problems that can be {\it efficiently solved using convolutional networks trained with gradient-descent, but at the same time it is hard to learn using a polynomial-size fully-connected network}.  Following the intuition provided by the approximation theorem above it is natural to further conjecture that given appropriate data, {\it learning shared weights  with gradient descent is ``easy'' for hierarchically local networks. } These and related results (See Appendix~\ref{sec:Pilot}) suggest that the hierarchy of visual processing cortical areas with local receptive fields is genetically hardwired, whereas effective weight sharing is actually learned from translation invariant images during development of the organism.

\vspace{10pt}
In the same spirit, we believe that the results and conjectures discussed above  could help shed light into one of the greatest
puzzles that has emerged from the empirical field of deep learning, which is trying to explain the
{\it unreasonable effectiveness} of convolutional deep networks in a
number of sensory problems. CNNs are indeed, a special case
of the hierarchical networks of the theorems above  (See Figure~\ref{fig:Network_Architectures}).

In this paper, we attempt to go from the mathematical why of the theorems to answering the questions of which are the visual tasks in which we should expect  an out-performance by convolutional networks  and which other visual tasks should be equally difficult for convolutional \textit{vs} fully connected networks. We will show that  the property of hierarchical compositionality of the target function  to be learned depends on $g(x)$, that is  on both the data ($x$) and the task ($g$). The approximation properties of the network $f$ should be matched to the task  ~\cite{zamir2018taskonomy,wang2019neural,conwellinterpreting,dwivedi2020unveiling,kunhardt2021effects}.

We will show evidence for this claim by showing a dissociation between different visual tasks (object recognition, texture perception and color estimation) on the same type of images as we disrupt the image locality prior through ``deterministic'' image scrambling. We find that systems in general will learn better a certain task when the approximation function $f$ matches the graph of  of the visual task $g$ as closely as possible.

\begin{figure*}[t!]
\centering
\includegraphics[width=1.0\linewidth,clip]{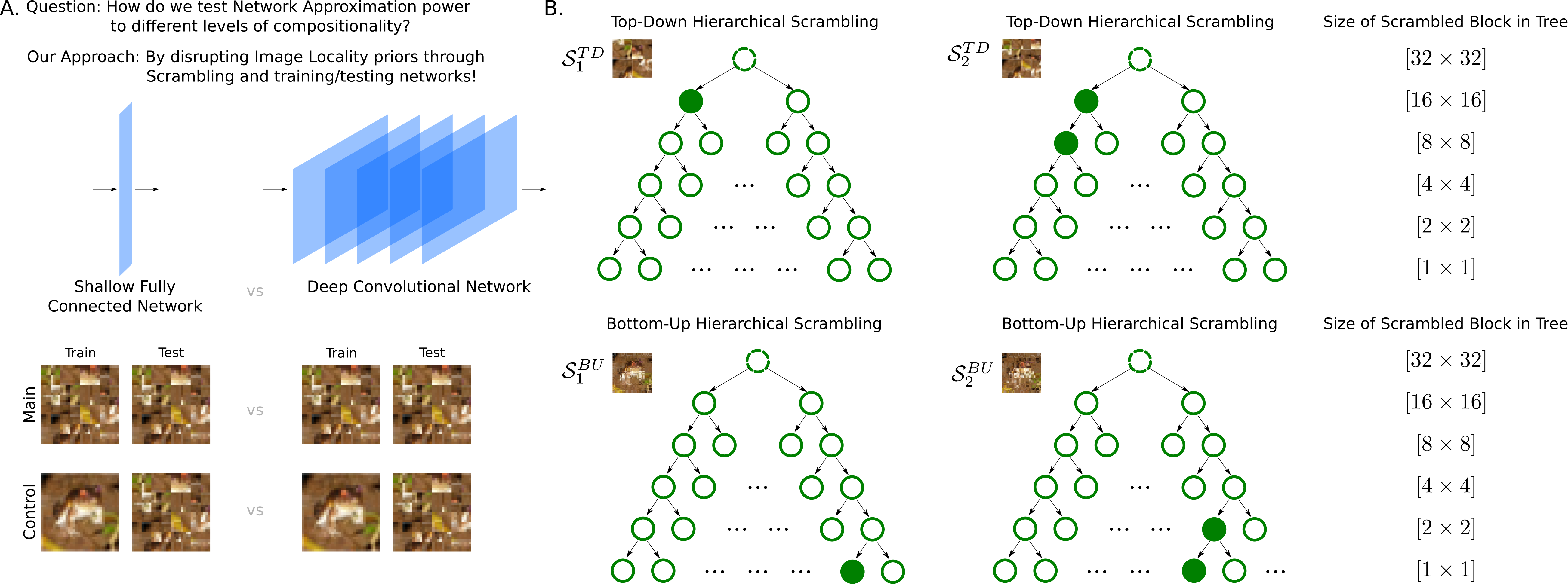}
\caption{\underline{A.} Our overall question is to analyze the contribution of compositional structure into networks that have a locality inductive bias such as deep convolutional neural networks \textit{vs} those that do not such as shallow fully connected networks (kernel machines). To do this, we will train/test different flavors of networks such as the before-mentioned on a variety of visual tasks where we will disrupt the locality of the data at different levels. \underline{B.} An example of how locality can be disrupted in a CIFAR image. Here we show different hierarchical scrambling operators $\mathcal{S}$ that follow a binary tree-like structure: top-down or bottom-up, to potentially explore how different networks will perform when trained/tested on these conditions.
}
\label{fig:experimental_setup}
\end{figure*}


\section{Hierarchical Compositional Tasks in Vision}

Before investigating the approximation power of different networks to a family of hierarchical compositional tasks, we will first define such tasks mathematically to check their compositional nature.

\subsection{Color Estimation}
A classical example of a visual task that is quite simple and requires no hierarchical structure is the color approximation task~\cite{emery2019individual,singh2020assessing,taylor2020representation,taylor2020representation,harris2019spatial,harris2020convolutional}. Indeed, color approximation, hereto the `average color' $C_I$ of an image $I$ can be computed exactly via the averaging function:
\begin{equation}
C_I = (1/N)\sum_i x_i
\end{equation}
where $N$ is the total number of $i$ pixels in the image (assuming a single channel), and where $x_i$ is the color intensity of the pixel at the $i$-th position.

If is easy to verify that a fully connected network of the shape $f(W;x) = W^Tx$ for a given set of weights $W$, can approximate $C_I$ exactly if $W$ is a column vector with each component $W_i=1/N$ for a vectorized image $x\in\mathbb{R}^d$. In fact, this function can also be approximated without a non-linearity as the usual structure of $f(\circ)$ generally yields the shape $f=\phi(x)=\sigma(Wx)$, for some $\sigma(\circ)$ non-linearity such as ReLU. Indeed, in this case the ReLU non-linearity would be redundant (as color is a positive value) and we can summarize the general form of color estimation as:
\begin{equation}
C_I=\phi(x)
\end{equation}
In classical views of vision science, one could classify color estimation (here computing the average color), as a low-level visual task~\cite{taylor2020representation}, as this task does not require higher-level cognitive processes such as identifying if it is an object/scene, or any sort of recognition-like processes. Thus in the context of this paper, we will define this low-level task as a global and non-compositional function or ``order 0'' Hierarchically Compositional, as the simplest form of computation of the function requires \textit{no compositionality} (nor locality) and can be approximated through a linear function. 

\subsection{Texture Perception}

Another visual task that increases in complexity given its compositional nature is the task of texture perception. Texture that is loosely viewed as ``stuff, not things"~\cite{heitz2008learning,balas2009summary,caesar2018coco,rosenholtz2014texture,rosenholtz2016capabilities,long2018mid}, where visual information is periodic at some level, can be better understood when using a generative (and parametric) texture modelling framework. Classic examples of these models include the Portilla~\&~Simoncelli~\cite{portilla2000parametric} texture synthesis model, that strongly influenced the Gatys~\textit{et~al.}~\cite{gatys2015texture} texture synthesis model -- which was the stepping stone to the pioneering field of Style Transfer~\cite{gatys2015neural} -- as such generative models reveal how textures can be rendered by following a mathematical formulation. In summary, these two pioneering approaches at heart consist of rendering a texture through the matching of second-order image statistics. These can be derived as a collection of cross-correlation of image transform outputs (\textit{i.e.} local feature maps $\phi(\circ)$), whether engineered through a Steerable Pyramid decomposition~\cite{simoncelli1995steerable} as in Portilla and Simoncelli, or learned through an optimization procedure as done with the multiple feature maps of a VGG19~\cite{simonyan2014very} that define a family of ``Texture Gramians'' as in the Gatys~\textit{et~al.} model~\cite{gatys2015texture}.

More formally these models define a Texture Matrix $T_M$ of a given image $x$ of the general form:
\begin{equation}
T_M = \begin{bmatrix}\phi_1(x) \phi_1(x) & \phi_1(x) \phi_2(x) & \dots & \phi_1(x) \phi_n(x)\\
\phi_2(x) \phi_1(x) & \phi_2(x) \phi_2(x) & \dots & \phi_2(x) \phi_n(x)\\
\vdots & \vdots & \ddots & \vdots\\
\phi_n(x) \phi_1(x) & \phi_n(x) \phi_2(x) & \dots & \phi_n(x) \phi_n(x)
\end{bmatrix}
\end{equation}
From here we can vectorize $T_M$, which will recast the same information (of form $\mathbb{R}^n\times\mathbb{R}^n$) as a column feature vector $T_F=[\{\phi_i(x),\phi_j(x)\}],\forall~i,j\in\mathbb{R}^T$. Naturally as texture perception can be defined as a classification task (\textit{i.e.} ``this texture belongs to this family of textures, but not to this other''~\cite{ziemba2016selectivity,freeman2013functional}) , we must have another hyper-plane decision function, which will recombine the feature vector products and also act on top of $T_F$ to produce an output from the collection of texture features. This yields an order-1 hierarchically local compositional function of the general form:
\begin{equation}
\label{eq:texure}
    T_P = \psi([\{\phi_i(x)\phi_j(x)\}]),\forall i,j
\end{equation}
Notice that in Equation~\ref{eq:texure}, there is already a notion of compositionality that by nature of the inner product of feature vector maps is non-linear. Thus texture is loosely being computed as \textit{``a function of a set of local functions of the image''} -- an idea that goes back to Julesz~\cite{julesz1981textons}, hence our attribution to giving texture the name of order-1 hierarchical compositionality.

\subsection{Object Recognition}

Perhaps the most challenging of all visual tasks is object recognition, which has been characterized as a highly compositional process -- dating back to works in perceptual psychology from Biederman~\cite{biederman1987recognition}, to the ever-more recent work of Hinton~\cite{hinton1988representing,hinton2021represent}. Indeed, for humans and machines to achieve robust object detection mechanisms, these systems must learn to recognize the object against a variety of conditions (or noise) that may affect the percept such as size, occlusion, illumination, viewpoint and clutter as suggested by Poggio~\cite{riesenhuber2000models,serre2005object,serre2007feedforward,poggio2011computational} and DiCarlo~\cite{dicarlo2007untangling,dicarlo2012does,pinto2008real}. 

Such theories of the representations of objects seem to suggest that objects can be identified as \textit{``a combination of parts, which are a combination of textures, which are a combination of edges, etc ...''} -- reminiscing to our definition of texture, though taken to a higher-order (or degree). This type of reasoning has both been posited in perception by Peirce~\cite{peirce2015understanding} when studying mid-level vision, and also in the Texform theory of Long~\&~Konkle~\cite{long2017mid,long2018mid,deza2019accelerated}, who found that objects coded as local texture ensembles can give rise to super-ordinate object categorization such as animacy without possibly encoding object identity. Similarly, computational approaches that support the higher-order compositional theory of objects have been found in the pioneering work of feature visualization of Zeiler~\&~Fergus~\cite{zeiler2014visualizing} and the feature invertibility models of Mahendran~\&~Vedaldi~\cite{mahendran2015understanding}. While it is true that deep networks can be biased to learn texture representations over shape for object categorization (see Geirhos~\textit{et~al.}~\cite{geirhos2018imagenet}), this problem can still be overcome with different data-augmentation strategies (see Hermann~\textit{et~al.}~\cite{hermann2019origins}), suggesting -- in a perhaps chicken-and-egg fashion -- that under the right conditions, the compositional nature of objects in hierarchical functions can indeed be learned.

We can thus ``generally'' pose the object recognition $O_R$ functional form as a cascade of localized non-linear functions $\phi_{\square}(\circ)$:
\begin{equation}
\label{eq:Object}
    O_R = \psi(\phi_n(\phi_{n-1}(...\phi_2(\phi_1(x),\phi'_1(x)))))
\end{equation}
While the formulation of Equation~\ref{eq:Object} is encouraging and perhaps even tractable for a purely feed-forward perceptual system, the non-obvious challenge is to find the correct family of non-linear local functions $\phi_{\square}(\circ)$, -- and \textit{how} and \textit{why} they change in the visual hierarchy~\cite{lindsey2019unified,harris2020anatomically} -- as this has been the main challenge of both human and machine vision since their conception.

\begin{figure*}[!t]
\centering
\includegraphics[width=1.0\linewidth,clip]{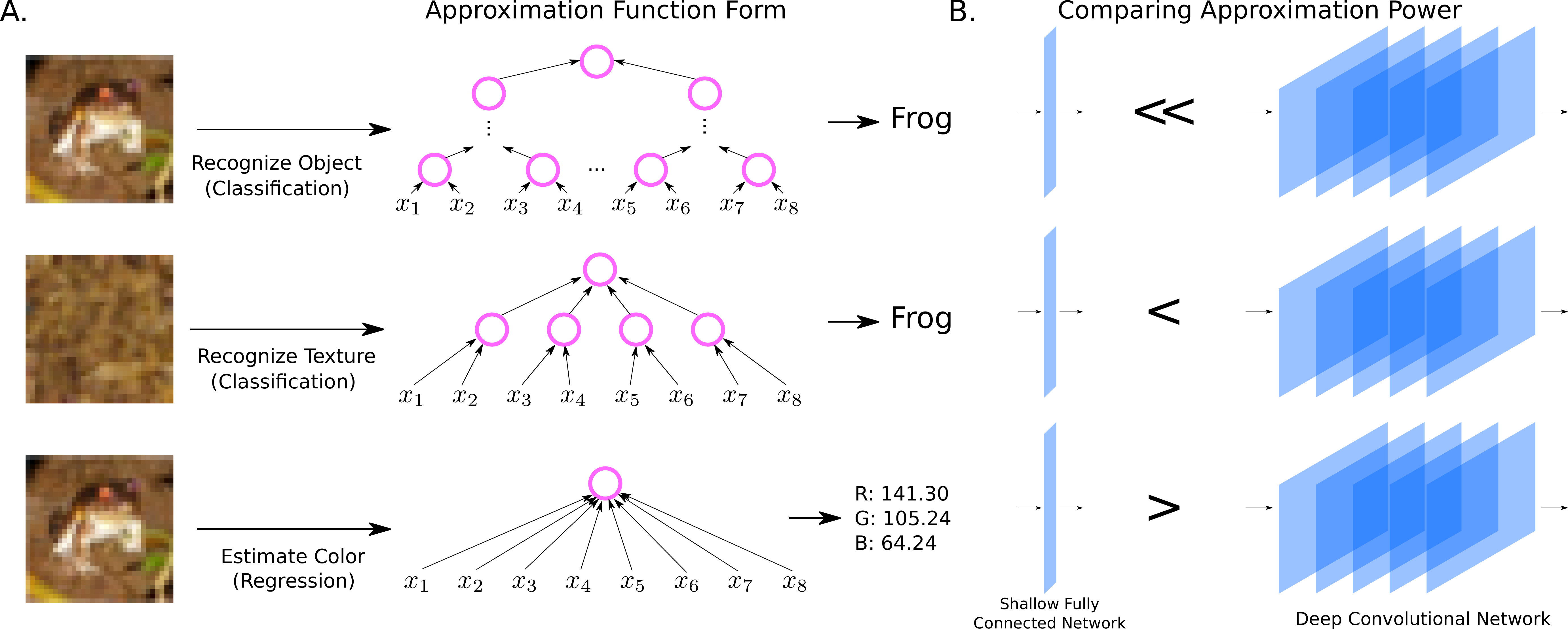}
\caption{\underline{(A.)}: A cartoon depicting the task manipulation to compare the approximation power between fully connected networks and deep convolutional networks given the compositional structure of the approximation task. Notice that although both tasks are different, the image distribution is the same (CIFAR-10). \underline{(B.)}: Top: If fully connected networks do indeed approximate trivially local tasks such as Color Estimation, then they will yield lower mean square error than deep convolutional networks. Middle: the case for Texture Perception that is compositional but to the first degree should show small advantages for convolutional networks over fully connected ones. Bottom: Similarly, if deep convolutional networks will exploit locality better than shallow fully connected networks, then they will achieve higher performance in the Object Recognition task -- and these perceptual gains should go away when trained/tested on \textit{shuffled images} where the locality prior is disrupted.}
\label{fig:Task_Manipulation}
\end{figure*}


\section{Methods}
\label{sec:Methods}

To investigate the role of the compositional structure in images on the approximation power of different neural networks, we must design a controlled experiment where we can manipulate their `degree' of compositionality. To accomplish this, we will hierarchically scramble the family of images over which the networks are trained/tested in a tree-like manner such that we can study the coarse-to-fine and fine-to-coarse implications of scrambling.

\vspace{-10pt}
\subsection{Image Scrambling} 
\vspace{-5pt}
Figure~\ref{fig:experimental_setup} shows a summary of our experimental framework. To test if the approximation power of one network type is greater than the other (\textit{e.g.} shallow non-convolutional vs deep convolutional), we will have them perform an approximation task for the unscrambled and scrambled conditions (\underline{A.}). To plot a performance curve as a discretized version of `level of scrambling', we propose two scrambling schemes: Top-Down scrambling where the image is scrambled in a coarse-to-fine manner, their dual form: Bottom-Up (fine-to-coarse) scrambling where the coarse image structure is preserved but the local components of the image are scrambled (\underline{B.} bottom). It is worth noting that the end-points of the scrambling tree (the fully scrambled top-down and bottom-up functions) are the same.

For all experiments we will show the i.i.d. performance of several networks when trained \& tested on the \textit{same} degree of scrambling (solid lines), and we will also as a reference an o.o.d scenario where networks are trained on unshuffled images and tested on shuffled images (dashed lines).

\vspace{-5pt}
\subsection{Dataset and Networks}
In what follows in the paper we will present our main set of results on experiments derived from CIFAR-10 objects. 
Our two main networks to be compared are a shallow fully connected network (SFCN) which consists of a single 10'000 hidden layer fully connected network with a ReLU non-linearity, and 5-stage deep convolutional network (DCN) with an additional 2-stage fully connected decision layer which is the classically known VGG11~\cite{simonyan2014very}. 

In addition to these two main networks we have 3 additional controls which are a ``Wide-Net'' which is a 2-layer wide shallow fully connected network that has that same number of parameters as the VGG11, a ``Deep-Net'' which is a 7-layer deep fully connected network that also has the same number of parameters as the VGG11, and a hyper small 6000 under-parametrized three-layer convolutional network which we call ``ThreeConvNet''. The goal of these controls will be to pin-point what the number of parameters is buying over the type of computation (convolution vs non-convolutional operators) and also to test if even an under-parametrized system with a convolutional prior can perform as well as a over-parameterized non-convolutional system.

All error bars shown in the following sections of the paper represent the standard deviation over 5 runs with different random weight initialization. Additional details with regards to the training and testing procedure and hyper-parameters for each experiment can be seen in the Supplementary Material.

\begin{figure*}[!t]
\centering
\includegraphics[width=1.0\linewidth,clip]{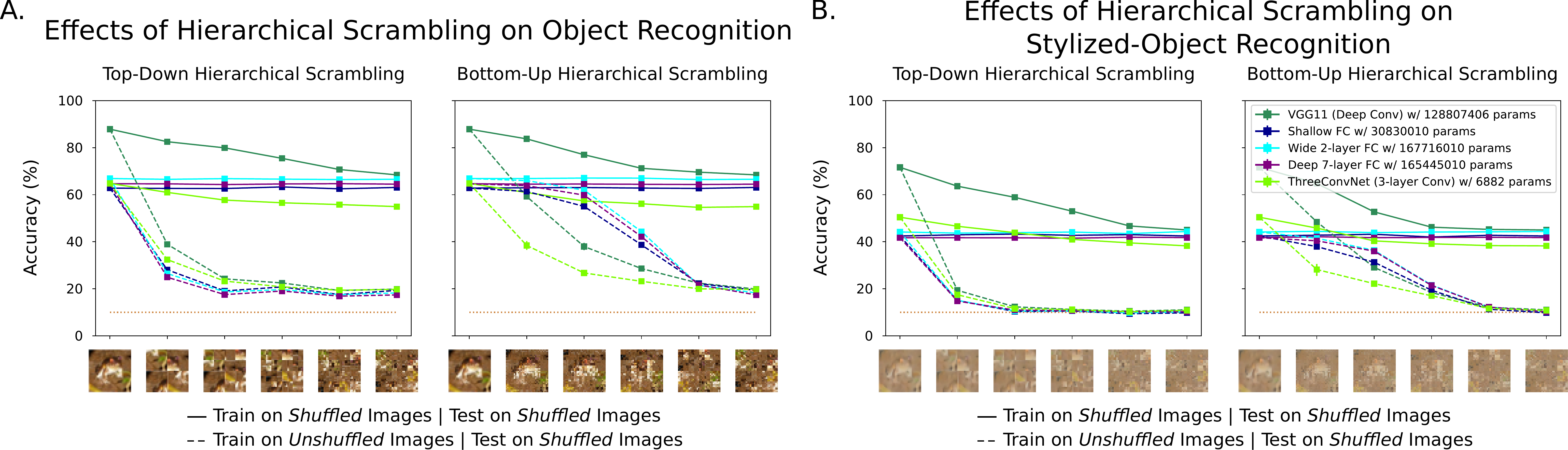}
\caption{\underline{A.} Results empirically showing that \textit{only} deep convolutional networks can exploit image locality priors for the hierarchically compositional task of object recognition. \underline{B.} Additional results showing that when the texture-bias is removed, that convolutional networks still have an exponential advantage in learning object recognition (a hierarchically compositional task) over non-convolutional networks (\textit{e.g.} see ThreeConvNet).}
\label{fig:Scrambling_Objects}
\end{figure*}

\section{Experiments}

In the following section we will go over our set of results in descending order of hierarchical compositionality, starting with the most complex task: object recognition, following with texture perception (compositional + local), and finally color estimation (that is neither a compositional or local task). A summary of our experiments can be seen in Figure~\ref{fig:Task_Manipulation} (A), accompanied by their approximation power hypothesis (B), where we conjecture that as the tasks become less hierarchical and compositional, that convolutional networks will lose their advantage, and that potentially fully connected networks will out-perform convolutional networks.

\subsection{Object Recognition: \normalfont{Deep Convolutional Networks can (1) approximate Hierarchically Local Compositional Tasks better than Shallow Fully Connected Networks AND (2) are \textit{affected} by scrambling}}

In our first experiment we will test to see how well do deep convolutional networks perform when doing a classical object recognition task and compare the performance to shallow fully connected networks. In particular, the goal of this experiment is to verify that DCN's will exploit the locality prior in the image structure and thus arrive to a higher performance than SFCN's which will not naturally converge to exploiting the local correlations in the image structure. In addition, we will evaluate how both of these networks behave as we train/test on different type of image scrambling manipulations where the locality prior is destroyed in different ways. As explained before in Section~\ref{sec:Methods}, the purpose of destroying the locality prior is to empirically verify that approximation functions such as those that have convolutional operators will \textit{not} excel as one would expect despite their over-parameterized nature.

Thus anticipating the differences in number of learnable parameters, we will also compare how the previously mentioned networks will perform to Wide-Net, Deep-Net and ThreeConvNet. The logic here is as follows: if even under matched amount of learning parameters, the deep convolutional network is the \textit{only} system that performs better for the unscrambled images, then we can claim that the benefit of DCNs in image recognition system is not due to their over-parameterized nature, but rather that they exploit image locality for a compositional task. Indeed, if this is the case, this benefit in performance should slowly disappear as we perform greater levels of scrambling on the images.

\underline{Results 4.1 A}: Figure~\ref{fig:Scrambling_Objects} (\underline{A.}) proves the previously mentioned hypothesis, where deep convolutional networks outperform the shallow fully connected network. The results with the additional controls are also overlayed, showing that the DCNS still outperforms the wide fully connected network (Wide-Net) and the deep fully connected network (Deep-Net) -- even when equalized with the same number of parameters. Naturally, the deep convolutional network outperforms ThreeConvNet as it is deeper despite both having a convolutional prior (more on this in Results 4.1 C).


\underline{Results 4.1 B}: In addition, we also notice as we increasingly disrupt locality through both the Top-Down and Bottom-Up variations of hierarchical scrambling, the difference in performance for the DCNs eventually goes away until the the image is fully scrambled at which performance is matched across networks that have the same number of learnables parameters (DCN, Wide-Net, Deep-Net). Furthermore, there is an asymmetry in the decay of performance for the dashed-lines (o.o.d testing: training on un-shuffled, testing on shuffled) Bottom-Up Hierarchical Scrambling operator where coarse information is roughly preserved. It would seem as if this last result implicitly verifies that fully connected networks are encoding long-range dependencies and coarse structure while deep convolutional networks are biased to encode localized feature correlations (\textit{i.e.} texture) as observed by Gatys~\textit{et~al.}~\cite{gatys2015texture}, Geirhos~\textit{et~al.}~\cite{geirhos2018imagenet} and Wendel~\&~Bethge~\cite{brendel2019approximating}, and recently Jacob~\textit{et~al.}~\cite{Jacob860759} in higher resolution images.

\underline{Results 4.1 C}: What is perhaps the most striking result is that the small under-parametrized 3-layer convolutional network with $\sim6000$ learnable parameters can still achieve roughly the same performance as all other over-parametrized fully connected networks, even those that exceed the number of parameters by an order of $10^5$ (Figure~\ref{fig:Scrambling_Objects} \underline{A.}). This last result complimentary proves the added benefit of convolutional operators on images with hierarchical structure for the compositional task of object recognition.

\vspace{-5pt}
\subsubsection{Stylized-CIFAR-10: Preservation of results when removing the texture-bias in CIFAR-10}
As an additional control with regards to the texture-bias phenomena~\cite{gatys2015neural,geirhos2018imagenet,brendel2019approximating}, we re-ran our experiments on a Stylized-CIFAR-10 dataset (following the procedure of Geirhos~\textit{et~al.}~\cite{geirhos2018imagenet} where we created a CIFAR-10 dataset with the texture of a collection of paintings similar to Stylized-ImageNet), and obtained the same pattern of results as in our previous experiments (See Figure~\ref{fig:Scrambling_Objects} \underline{B.}). These findings support the idea that the object recognition task is hierarchically compositional and does benefit exponentially from the convolutional prior when image locality is preserved and other cue-conflicting biases are removed.

What is also worth noting is that the under-parametrized ThreeConvNet now shines and outperforms all other over-parameterized fully connected networks (wide or shallow or deep). This last results further shows the computational tractability and relative ease of convolutional neural networks to exploit localized hierarchical structure in a compositional manner when all other biases are removed~\cite{geirhos2020shortcut}.

\subsection{Texture Perception: \normalfont{Deep Convolutional Networks better approximate Local Compositional Tasks than Shallow Fully Connected Networks and are \textit{un-affected} by scrambling}}

The next task we decided to explore was a lower level hierarchically compositional task -- a task of texture perception, as it is known that texture perception in both humans and machines can be modelled via second-order image statistics~\cite{julesz1981textons,portilla2000parametric,gatys2015texture,deza2018towards,wallis2019image,vacher2020texture,herrera2021flexible}. The peculiarity about this task is that it is still hierarchical but not as hierarchical as object recognition that goes from edges to shapes: $(edges \rightarrow textures \rightarrow parts \rightarrow shapes)$. This contrived control condition allows us to examine at a more nuanced level what can convolutional structure provide (if anything) when the approximation task is not \textit{too} compositional.

\begin{figure}[!t]
\centering
\includegraphics[width=1.0\linewidth,clip]{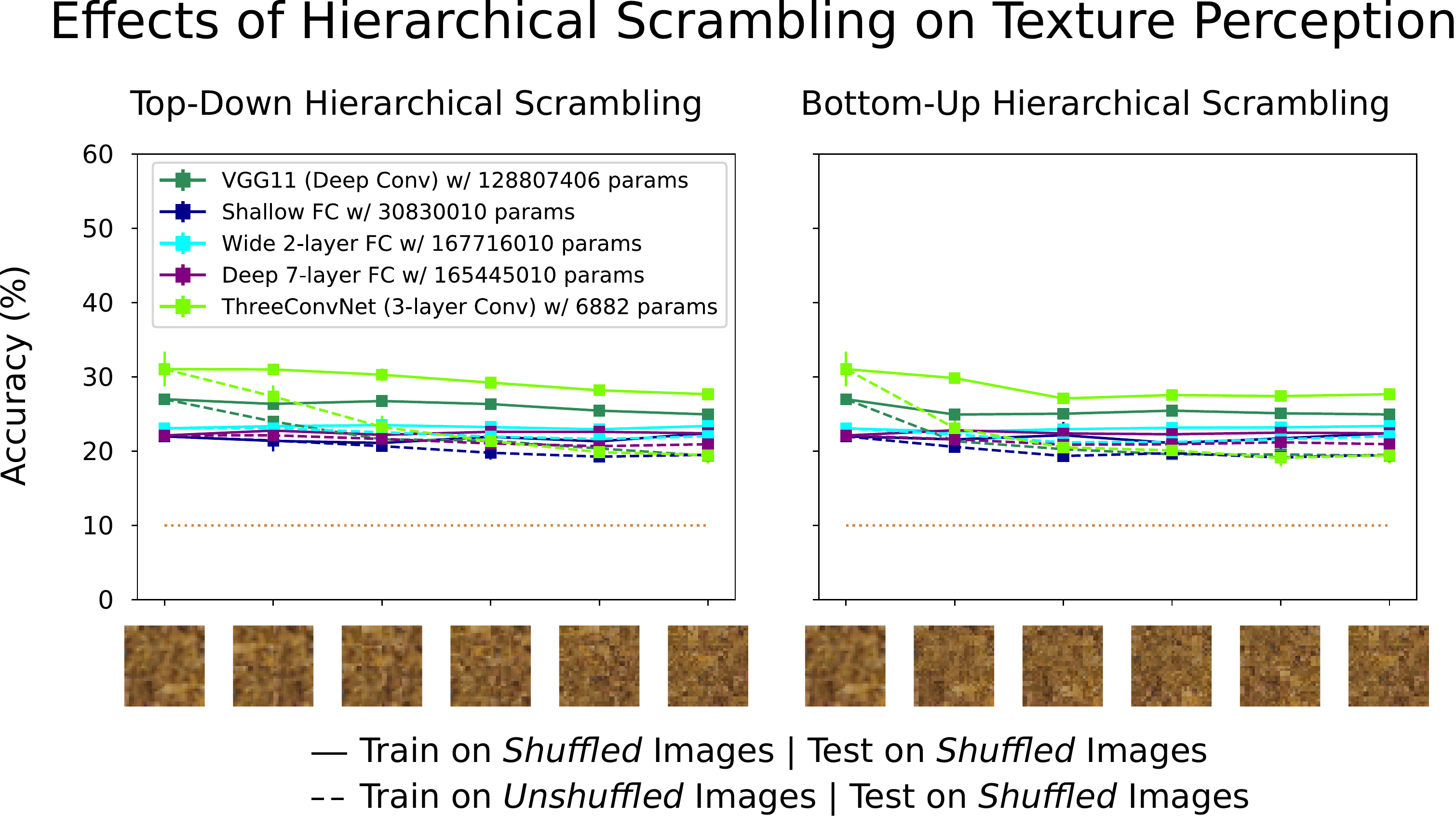}
\caption{Results showing that convolutional networks better approximate texture than fully connected networks -- as there is a small (non-obvious) advantage in exploiting locality (Appendix~\ref{sec:App_Feature_Vis}). Even \textit{under-parameterized} convolutional networks approximate texture better than all other networks. What is more surprising is that the \textit{approximation power of all networks is unaffected by disrupting the locality priors through image scrambling} -- also suggesting that CNNs maintain performance via hierarchical computation.
}
\label{fig:Scrambling_Texture}
\vspace{-5pt}
\end{figure}

To do this, we rendered a texturized-CIFAR dataset by running a set of normally distributed noise images that were colored with their corresponding CIFAR-10 image through ZCA, to later run these noise images through a Style Transfer operator (Adaptive Instance Normalization~\cite{huang2017arbitrary}) thus rendering an equivalent texture to the original CIFAR-10 image -- an approach loosely inspired from Deza~\textit{et~al.}~\cite{deza2018towards} or the Texforms from Long~\textit{et~al.}~\cite{long2018mid,deza2019accelerated}. Altogether, this pipeline rendered 50000 small textured images corresponding to each CIFAR-10 image with their matches classes. The task these networks are then expected to learn via a standard cross-entropy loss is a `texturized-object' categorization task where the matching class is the category from which the texturized CIFAR-10 image was rendered.

\underline{Results 4.2 A}: Figure~\ref{fig:Scrambling_Texture} shows that deep convolutional networks slightly overperform shallow fully connected networks on this order-1 hierarchically compositional task. However, a closer look at the previous figure shows us that even when training and testing on the same type of scrambled textures, both deep convolutional networks and shallow fully connected networks are un-affected by this manipulation as the solid lines stay mostly horizontal. While perhaps counter-intuitive, this last result matches the expected theoretical outcome for the deep convolutional network as it learns to capitalize on local structure (even with overlapping receptive fields) to learn these Texture Gramian-like structures from Gatys~\textit{et~al.}~\cite{gatys2015texture}.

\underline{Results 4.2 B}: We then examined how the additional control networks would perform (also Figure~\ref{fig:Scrambling_Texture}) and found a similar pattern where both convolutional networks perform better than all other fully connected networks, including the heavily under-parameterized ThreeConvNet that even surpasses the performance of the VGG11. It is possible that the under-parametrization has aided as an advantage to not overfit and correctly represent the aggregate texture of each object class. It is also possible that this advantage has come through via another factor: kernel size, given that ThreeConvNet has kernels with greater receptive field size ($5\times5$ in ThreeConvNet \textit{vs} $3\times3$ in VGG11; Appendix~\ref{sec:App_Feature_Vis}.).

\vspace{-5pt}
\subsection{Color Estimation: \normalfont{Shallow Fully Connected Networks better approximate Global and non-Compositional Tasks than Deep Convolutional Networks}}

Finally, our last experiment is a color estimation experiment to investigate how well both deep convolutional networks and fully connected can estimate the normalized color of an image. To do this, we trained both networks to output a 3-dimensional column real-valued vector with normalized Red, Green and Blue channel information that was the average global color value under a Mean Square Error (MSE) loss. If the hypothesis of proper ``function-to-task'' matching is correct, then shallow fully connected networks should achieve greater performance (\textit{i.e. less error}) that deep convolutional networks when computing this value over a set of testing images. We used the full collection of 50000 un-modified CIFAR images for training and the remaining 10000 images for testing for this color approximation task, where the image class is irrelevant.

\begin{figure}[!t]
\centering
\includegraphics[width=1.0\linewidth,clip]{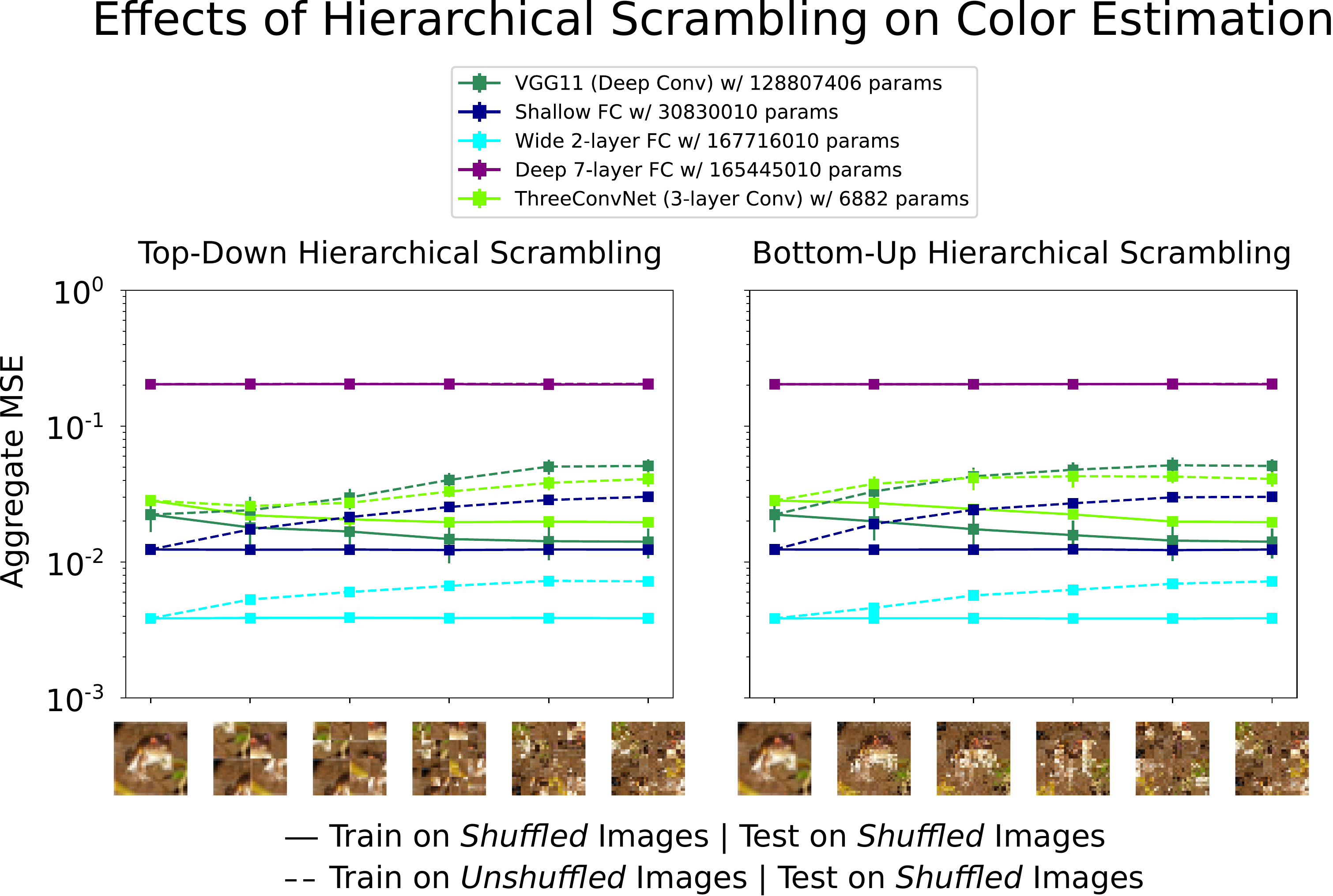}
\caption{Results showing that shallow fully connected networks achieve less approximation error that deep convolutional networks when computing the average color of an image -- a global task that requires no hierarchical function. When fixing the number of parameters, and varying hierarchical structure, shallow fully connected layers continue to achieve smaller error, while deep fully connected networks perform worse (Wide-Net vs Deep-Net). These set of results \textit{show that convolutional + deeper is not always better} contingent on the approximation task.
}
\label{fig:Scrambling_Color}
\vspace{-8pt}
\end{figure}

\underline{Results 4.3 A}: Figure~\ref{fig:Scrambling_Color} shows that for the unscrambled condition, shallow fully connected network achieve reduced mean square error rates over deep convolutional networks for the ensemble of testing images. However, as the locality prior is gradually destroyed through scrambling, we notice that the deep convolutional network asymptotes to the performance of the SFCN (lower error is better). These results suggest that the convolutional operator still tries to exploit the locality prior to search for compositional structure even if the approximation function is non-hierarchical.

\underline{Results 4.3 B}: Figure~\ref{fig:Scrambling_Color} further proves the previous point by showing that a shallow and wider fully connected network with greater number of parameters (Wide-Net) yields a smaller error over all other models, while the most hierarchical of all models a 7-layer fully connected network (Deep-Net) achieves the highest error (performs worse). Similarly, we observe that the high performing texture-perception ThreeConvNet is now at a disadvantage against all other networks. This last observation indeed agrees with the theory, as color estimation is a global property and the convolutional prior of ThreeConvNet works at a disadvantage -- despite still being able to approximate color reasonably well.

\section{Discussion}


In this paper we reviewed the importance and critical cases of how and when deep convolutional networks succeed and exponentially do better than shallow fully connected networks (object recognition), when they have a minimal advantage (texture perception) -- but also counter-intuitively when they can do worse (color estimation) -- as modulated by the visual approximation task. 

There have also been several theoretical works that have also supported some of our experimental results. For example, Poggio~\text{et~al.}~\cite{poggio2020theoretical} 
shows that convolutional networks -- that exploit locality -- require an exponentially smaller number of parameters to generalize than fully connected networks. The ratio of the sample complexity is in the order of $\frac{N_{deep}}{N_{shallow}} \approx \epsilon^d$. Empirical work by Elsayed~\textit{et~al.}~\cite{elsayed2020revisiting} and Neyshabur~\cite{neyshabur2020towards} has shown the importance of the convolutional prior as an inductive bias and how certain regularization priors may induce learned locality in a fully connected neural network. Consistently with our experiments, Brendel~\&~Bethge~\cite{brendel2019approximating} have also shown that local pixel structure \textit{still} supports object recognition even if the higher-order compositionality of an object is destroyed (they scramble patches that are large for this results to hold).

Overall, this work reminds us to not loose sight on evaluating and understanding the tasks for which different perceptual systems work. With the advent of new models such as Transformers~\cite{dosovitskiy2021an,jaegle2021perceiver} it will continue to be important to not only study their robustness, generalization \& convergence properties but also to understand for what tasks do such models out-perform (or under-perform) classical neural network models and why.

\newpage

{\small
\bibliographystyle{ieee_fullname}
\bibliography{Locality}
}

\cleardoublepage

\begin{appendix}

\section{Training Schemes across All Networks}

All our experiments were performed with PyTorch 1.4.0, and our code will be made publicly available.

\subsection{Learning}

The set of networks we used were trained with SGD and the following learning rate $(\eta)$ scheduled by:

\begin{equation}
    \eta = 0.1\times \text{lr\_network\_adjustment} \times \text{dataset\_learning\_factor}
\end{equation}

where the learning rate was halved after 25\% of the total number of epochs per experiment, and halved again after reaching 50\% of the total number of epochs of the experiment. The SGD momentum was set to $0.9$, the weight decay was set to 0.0005, the dampening factor was set to $0$, and we used the Nesterov momentum update rule. In addition we had the following hyper-parameters:\\

\textbf{Total Number of Training Epochs}:\\
Object Recognition: 100 epochs \\
Stylized-Object Recognition: 100 epochs \\
Texture Perception: 100 epochs \\
Color Estimation: 40 epochs \\

\textbf{lr\_network\_adjustment}:\\
VGG11 (Deep Conv): 1.0\\
Shallow\_FC: 0.5\\
Wide-Net: 0.5\\
Deep-Net: 0.5\\
ThreeConvNet:  1.0\\

\textbf{dataset\_learning\_factor}:\\
Object Recognition: 0.1\\
Stylized-Object Recognition: 0.25\\
Texture Perception: 0.25\\
Color Estimation: 0.001\\

\textbf{Loss Function}:\\
Object Recognition: Cross-Entropy\\
Stylized-Object Recognition: Cross-Entropy\\
Texture Perception: Cross-Entropy\\
Color Estimation: Mean Square Error\\

\textbf{Batch Size}:\\
VGG11 (Deep Conv): 64\\
Shallow\_FC: 64\\
Wide-Net: 64\\
Deep-Net: 64\\
ThreeConvNet: 64\\

\subsection{Data-Augmentation}
All networks were trained with the following data-augmentation transform of : Random Cropping (area-ratio: 0.7 to 1.0), followed by re-sizing back to $32\times32$, followed by random horizontal flipping with a 0.5 chance of a flip, followed by color normalization with the RGB means set to: (0.485,0.456,0.406), and the standard deviation to (0.229,0.224,0.225). No data-augmentation was used for testing, except for the previously mentioned color normalization to preserve learned lower-level image statistics.

These hyper-parameters did not change as we varied the scrambling condition.

\subsection{Matched random weight initialization across Scrambling conditions}

An additional nuance about our training pipeline is that for each of the 5 independently sampled runs per network per experiment, the networks all started with \textit{the same} random weight initialization across all scrambling conditions. For example, for the for the run\_id $1$ of the VGG11 for Object Recognition there are a collection of 10 separately trained networks trained on the different scrambling procedures: $\{\mathcal{S}_0,\mathcal{S}_1^{TD},\mathcal{S}_2^{TD},\mathcal{S}_3^{TD},\mathcal{S}_4^{TD},\mathcal{S}_1^{BU},\mathcal{S}_2^{BU},\mathcal{S}_3^{BU},\mathcal{S}_4^{BU},\mathcal{S}_5\}$, all starting from the same point in weight space $\theta^{(1)}$, and converging to their own solutions: $\{\theta_0^{(1)},\theta_1^{(1),TD},\theta_2^{(1),TD},\theta_3^{(1),TD},\theta_4^{(1),TD},\theta_1^{(1),BU},\theta_2^{(1),BU}$,\\$\theta_3^{(1),BU},\theta_4^{(1),BU},\theta_5^{(1)}\}$. Similarly, for the 2nd run of experiments each network will start with $\theta^{(2)}$, and converge to their own solution space -- and so forth for each of the 5 independently sampled runs for each network per experiment. This choice was made to reduce the variance that could have affected the performance of one network for a particular scrambling condition over another scrambling condition purely driven by a ``lucky initialization''.

\subsection{Notes on Deterministic Scrambling}

\begin{figure*}[!t]
\centering
\includegraphics[width=0.8\linewidth,clip]{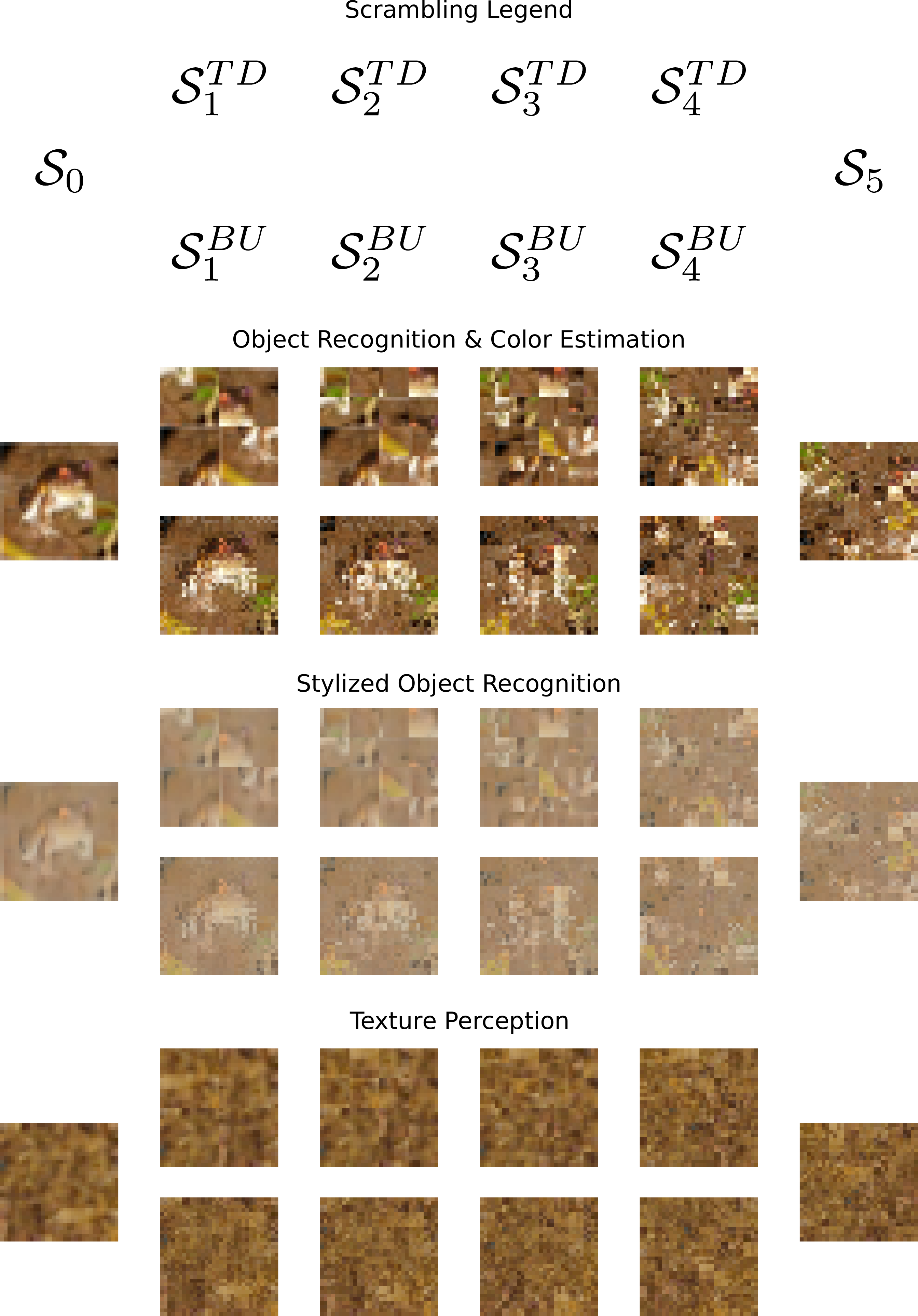}
\caption{A zoomed legend of the types of scrambling maps used in our experiments. Notice that the mappings resemble the hierarchical structure of a binary tree and they have a recursive flavor. These scrambling maps are preserved across networks, tasks \& runs in our experiments.}
\label{fig:Zoom_Scrambling}
\end{figure*}

It is worth noting that the scrambling map for a specific setting (\textit{e.g.} $\mathcal{S}_2^{TD}$ in Object Recognition for run id 3 in Shallow FC) is \textit{preserved} throughout training and testing. In other words if the random mapping moves pixel 1 to position 32 and pixel 2 to position 5, (etc...) at training; the \textit{same} mapping is done at testing. 
More specifically, scrambling maps are constrained by the scrambling procedure and are different as we vary across scrambling condition (\textit{e.g.} $\mathcal{S}_2^{TD}$ vs $\mathcal{S}_2^{BU}$), but held \textit{constant} across experiments (\textit{e.g.} Object Recognition vs Texture Perception), networks (\textit{e.g.} VGG11 vs Wide-Net), and run id (\textit{e.g.} run id 1 vs run id 3). Please see our Scrambling Legend for a better visualization of the exact scrambling maps used in our experiments in Figure~\ref{fig:Zoom_Scrambling}

\newpage
\cleardoublepage

\newpage
\cleardoublepage

\section{Loss Function Convergence across Multiple Experiments}

There are several observations to be made about the convergence of the loss for the set of 4 experiments presented in this paper.

\subsection{Object Recognition}

We find that when all hyper-parameters are held constant for the object recognition task, the VGG11 has an easier time reducing its training error when the data is un-scrambled ($\mathcal{S}_0$) \textit{vs}  when it is fully scrambled ($\mathcal{S}_5$), or in general, as the ``order of scrambling'' increases: $\mathcal{S}_1^{BU}\rightarrow\mathcal{S}_4^{BU}$ and $\mathcal{S}_1^{TD}\rightarrow\mathcal{S}_4^{TD}$. Naturally, an orthogonal observation can be made for the convergence of the loss for all the fully connected networks (Shallow FC, Deep-Net and Wide-Net), where scrambling the image does not (and should not) affect learning as these networks have no structural or locality priors.

We see very similar trends for Stylized Object Recognition, with the exception that both deep networks: VGG11 (convolutional) and Deep-Net (non-convolutional), behave strangely in the last 40 epochs of training (see mainly $\mathcal{S}_0$, $\mathcal{S}_1^{BU}$, $\mathcal{S}_1^{TD}$) -- despite still managing to succeed when generalizing to the unseen testing images. What is also quite surprising is that there seems to be minimal learning done on the ThreeConvNet (though the y-axis scale is logarithmic). Interestingly, it is the ThreeConvNet that out-performs all other fully connected networks -- the latter group of which manage to show better \& faster learning on the training set. Perhaps the reason for their shortage of success is that as they do not have a locality prior, then such networks are forced to pickup other potential cues (indirectly) about color, edge or curvature statistics, and yet does not manage to surpass the generalization ability of the ThreeConvNet.

\subsection{Texture Perception}

The texture perception task shows a high convergence of the training loss on the VGG11, Shallow FC, Deep-Net and Wide-Net architectures -- however these are the networks that do not perform well on the unseen testing images. It is (quite strangely) the case that ThreeConvNet achieves minimal learning on the training set, but has actually managed to generalize well and \textit{better} than all other networks to the unseen testing images (textures). This trend is apparent independent of the level of scrambling.

\subsection{Color Estimation}

The color estimation tasks shows us that color estimation is approximated faster with shallow networks (Shallow FC and Wide-Net) instead of deep or convolutional networks. The plots shown for the training loss convergence mirror the color estimation accuracy for the unseen testing images. There is no strong difference in convergence of such loss across networks, although our experimental results seem to suggest that VGG11 may converge faster to a better solution when the image is fully scrambled as it will force the network to \textit{not} pick up on any locality priors for such a global task.

\begin{figure*}[!t]
\centering
\includegraphics[width=0.65\linewidth,clip]{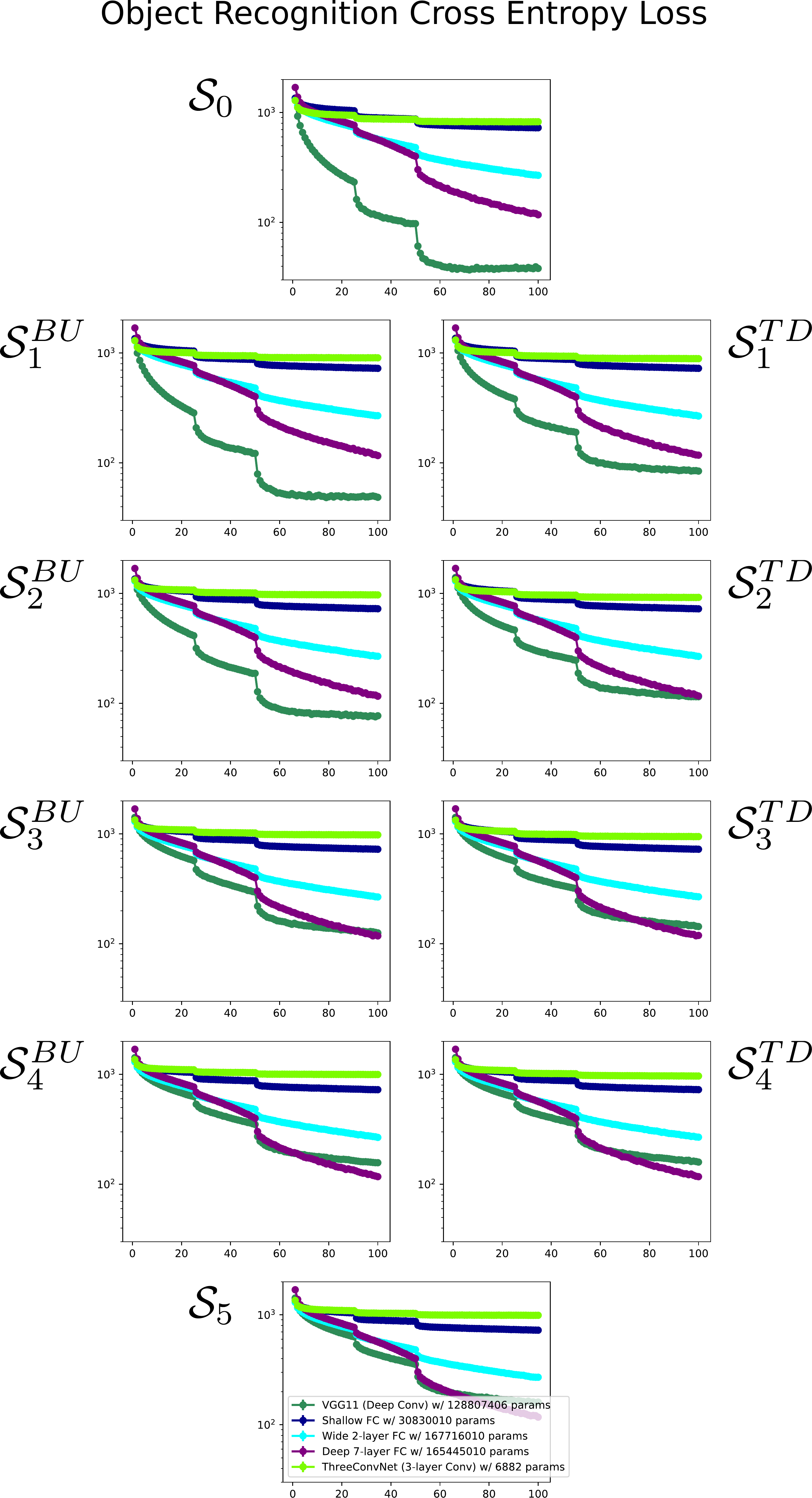}
\caption{Convergence of Loss Function value as a function of number of training epochs for the Object Recognition task across all networks and scrambling procedures. Errorbars denote standard deviation.}
\label{fig:Object_Recognition_Loss}
\end{figure*}

\newpage

\begin{figure*}[!h]
\centering
\includegraphics[width=0.65\linewidth,clip]{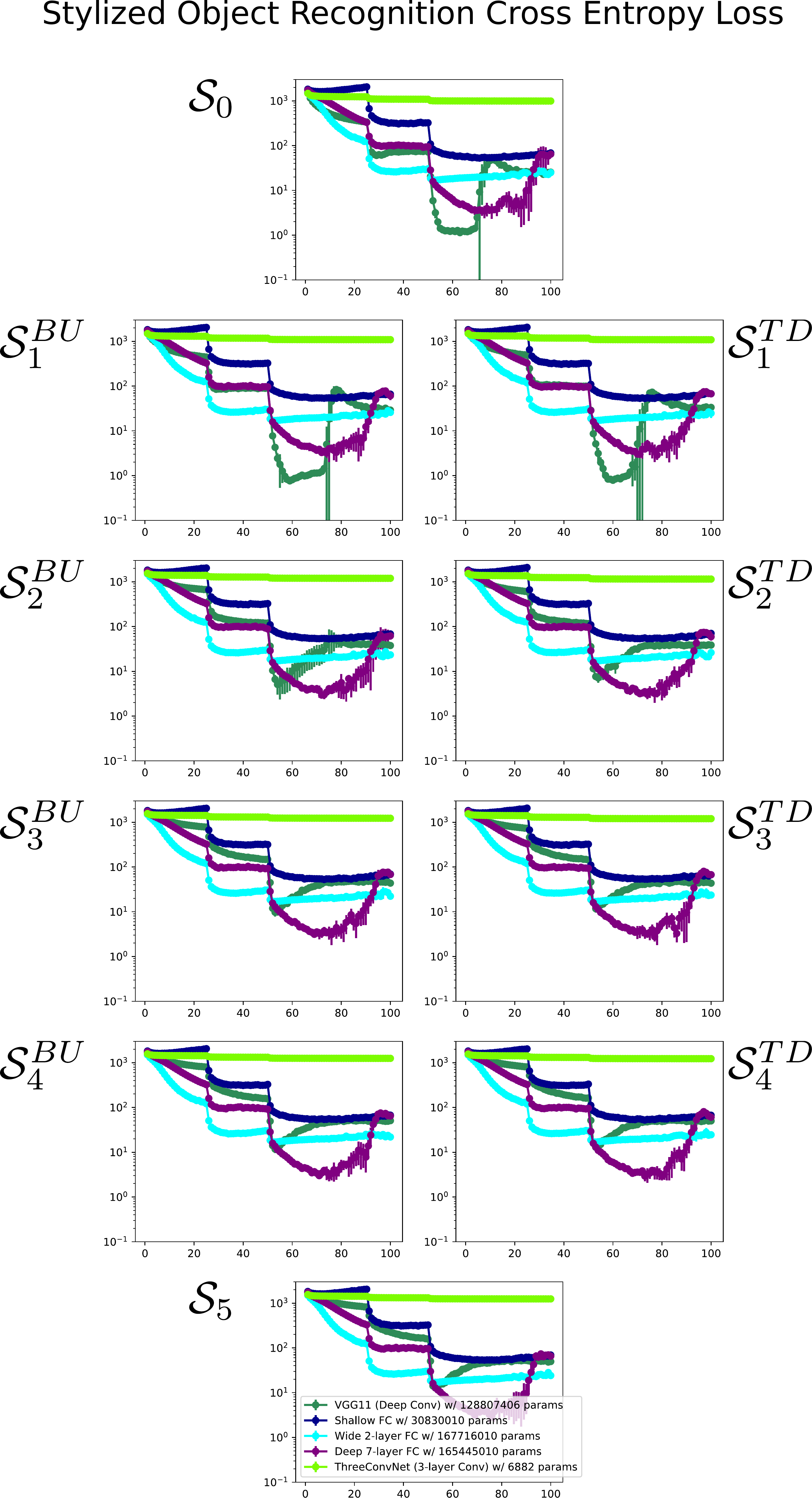}
\caption{Convergence of Loss Function value as a function of number of training epochs for the Stylized Object Recognition task across all networks and scrambling procedures. Errorbars denote standard deviation.}
\label{fig:Stylized_Object_Recognition_Loss}
\end{figure*}

\newpage

\begin{figure*}[!h]
\centering
\includegraphics[width=0.65\linewidth,clip]{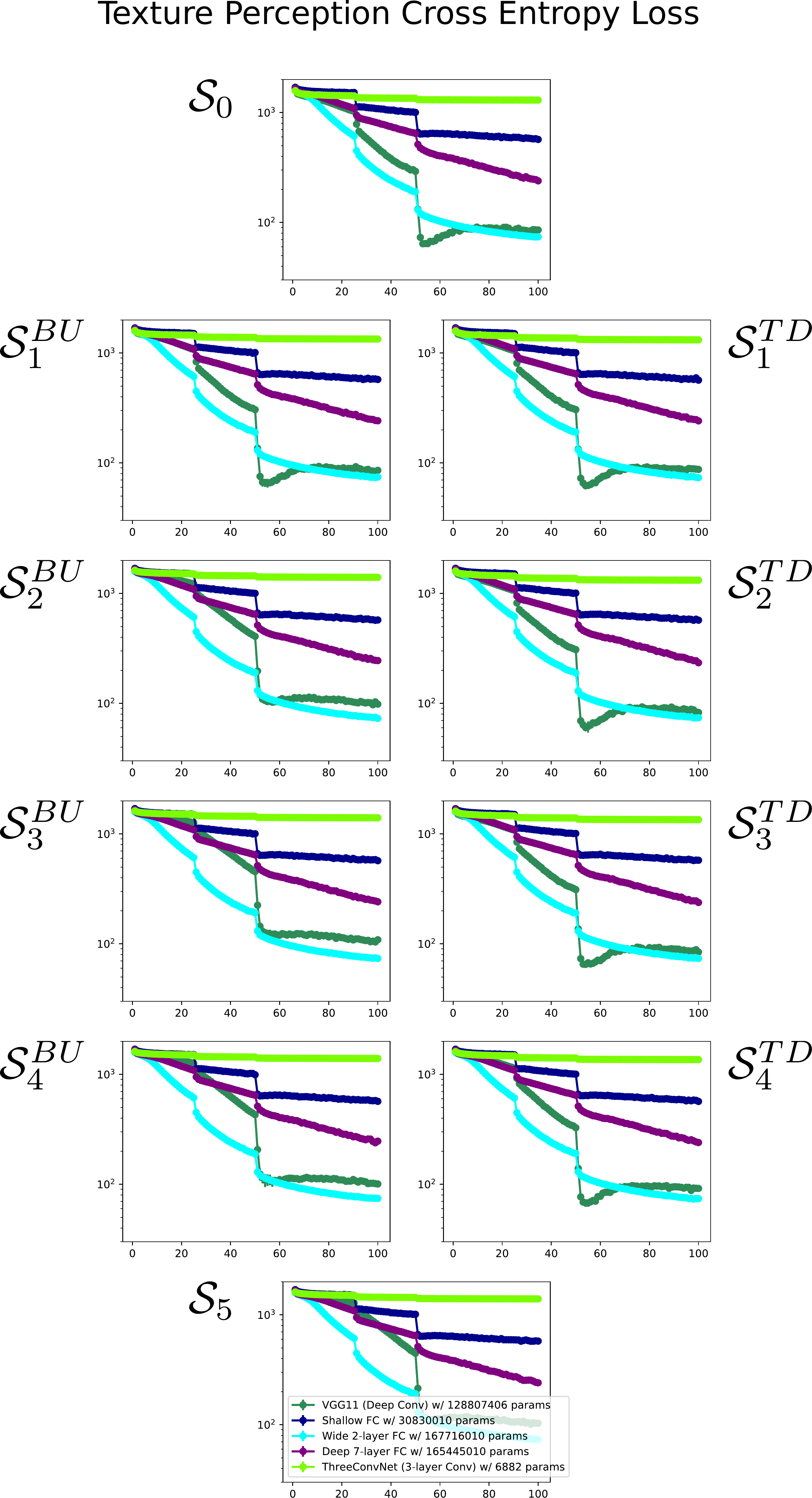}
\caption{Convergence of Loss Function value as a function of number of training epochs for the Texture Perception task across all networks and scrambling procedures. Errorbars denote standard deviation.}
\label{fig:Texture_Perception_Loss}
\end{figure*}

\newpage

\begin{figure*}[!h]
\centering
\includegraphics[width=0.65\linewidth,clip]{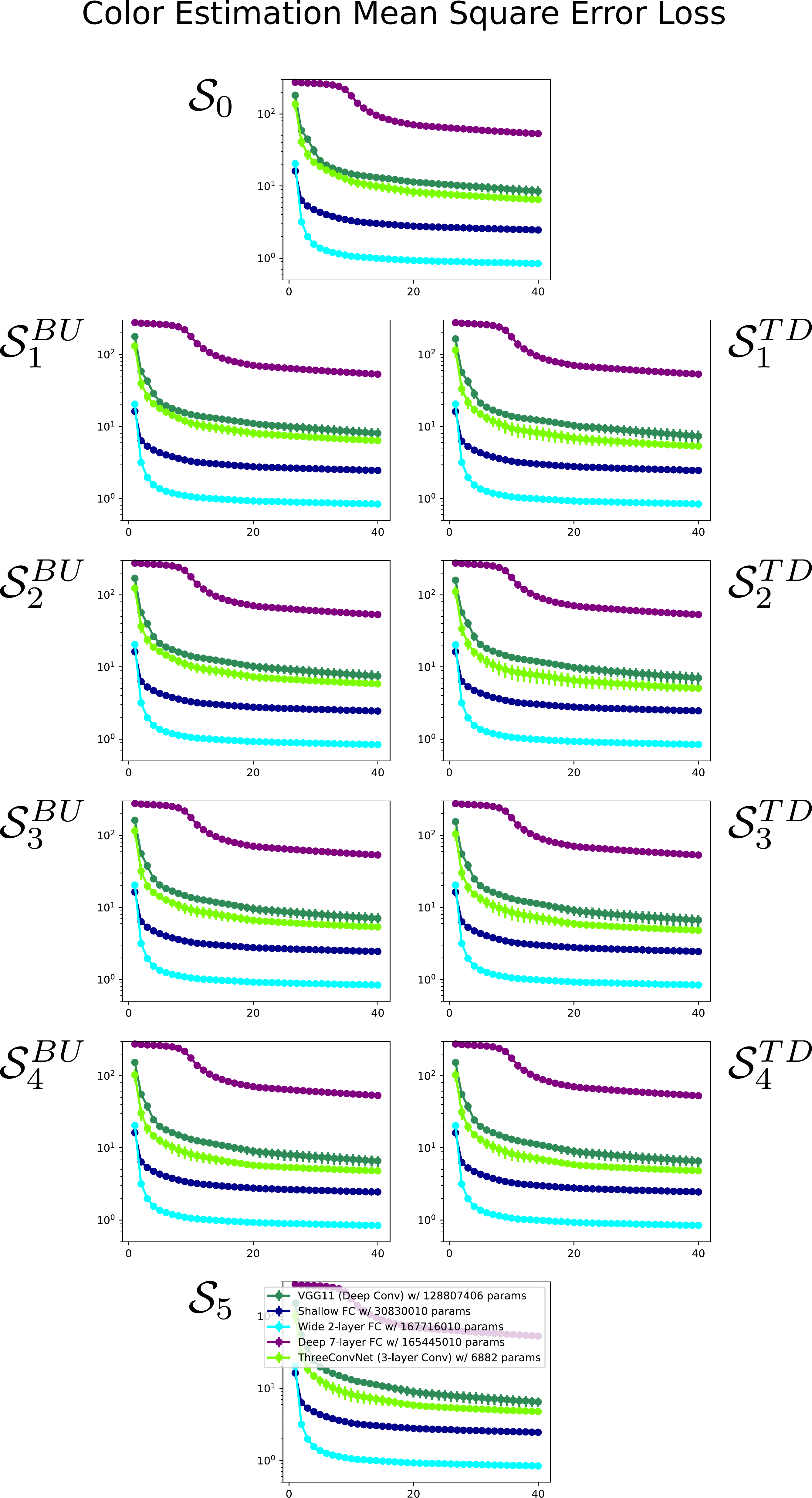}
\caption{Convergence of Loss Function value as a function of number of training epochs for the Texture Perception task across all networks and scrambling procedures. Errorbars denote standard deviation.}
\label{fig:Color_Estimation_Loss}
\end{figure*}

\cleardoublepage
\cleardoublepage
\newpage

\section{Feature Visualization of the Convolutional Networks in the Visual Tasks}
\label{sec:App_Feature_Vis}

In the following section we will show a collection of feature visualizations for the convolutional networks VGG11 and ThreeConvNet for each of the 4 experiments: Object Recognition, Stylized-Object Recognition, Texture Perception and Color Estimation for both the unscrambled $(\mathcal{S}_0)$ and fully scrambled condition $(\mathcal{S}_5)$. Some comments for each experiment and the respectively learned features can be seen in the rest of this section.

\subsection{ThreeConvNet}

\begin{figure*}[!h]
\centering
\includegraphics[width=0.8\linewidth,clip]{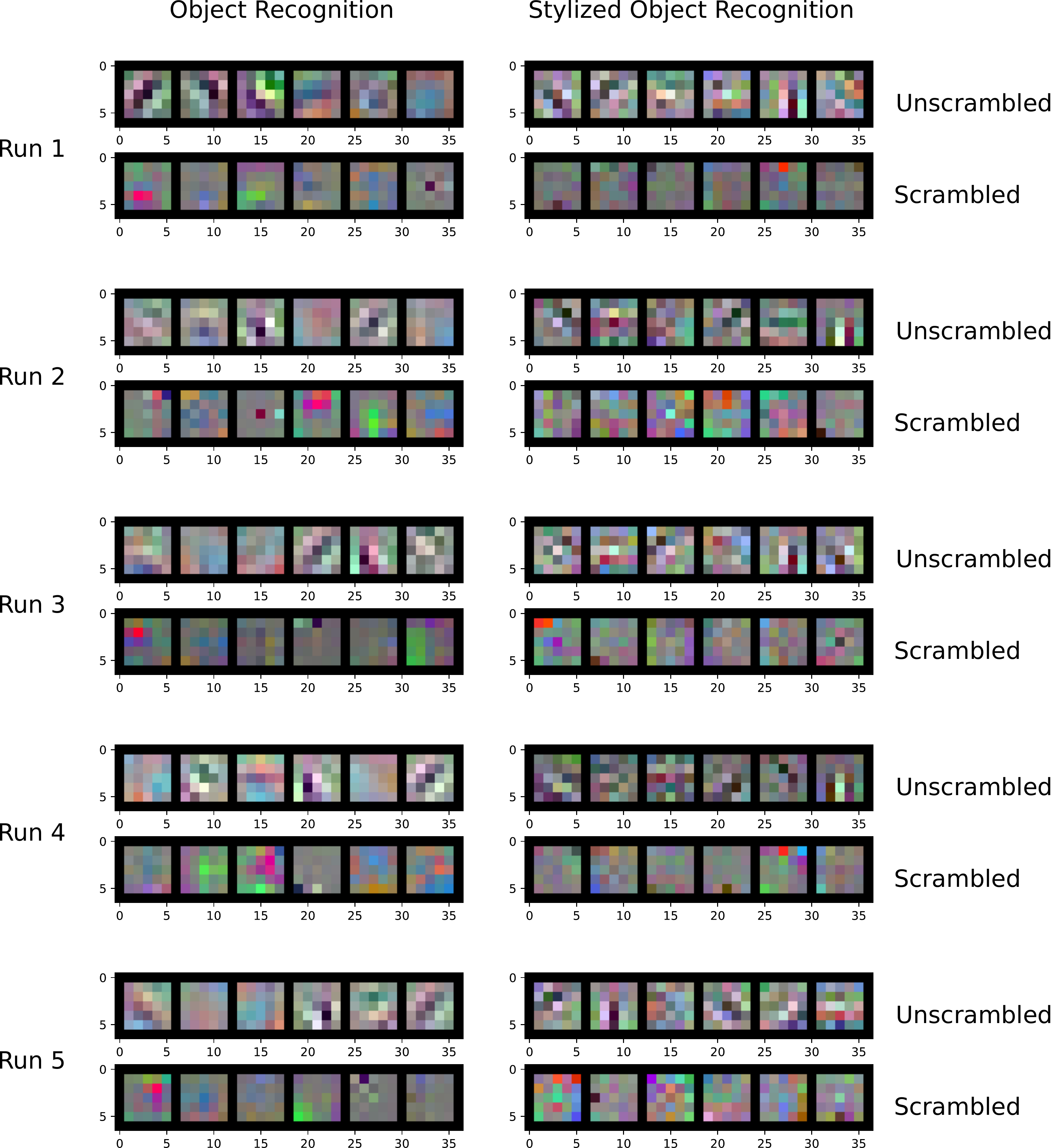}
\caption{ThreeConvNet filters for Object Recognition and Stylized Object Recognition.}
\label{fig:ThreeConvNet_A}
\end{figure*}

\textbf{Object Recognition}: Figure~\ref{fig:ThreeConvNet_A} shows that the ThreeConvNet features across all runs show strong diagonally-tuned ($45\deg$ or $135\deg$) oriented filters, with some level of color opponency under the unscrambled condition. The same can not be said for the fully scrambled condition as all local structure is disrupted.

\textbf{Stylized Object Recognition}: The same set of diagonally orientation tuned filters emerge for the Stylized Object Recognition task in the unscrambled condition (Figure~\ref{fig:ThreeConvNet_A}). This is expected given the removal of the texture bias. Naturally, there is no obvious pattern of the learned features for the scrambled condition -- similar to the object recognition experiment. 

\begin{figure*}[!h]
\centering
\includegraphics[width=0.8\linewidth,clip]{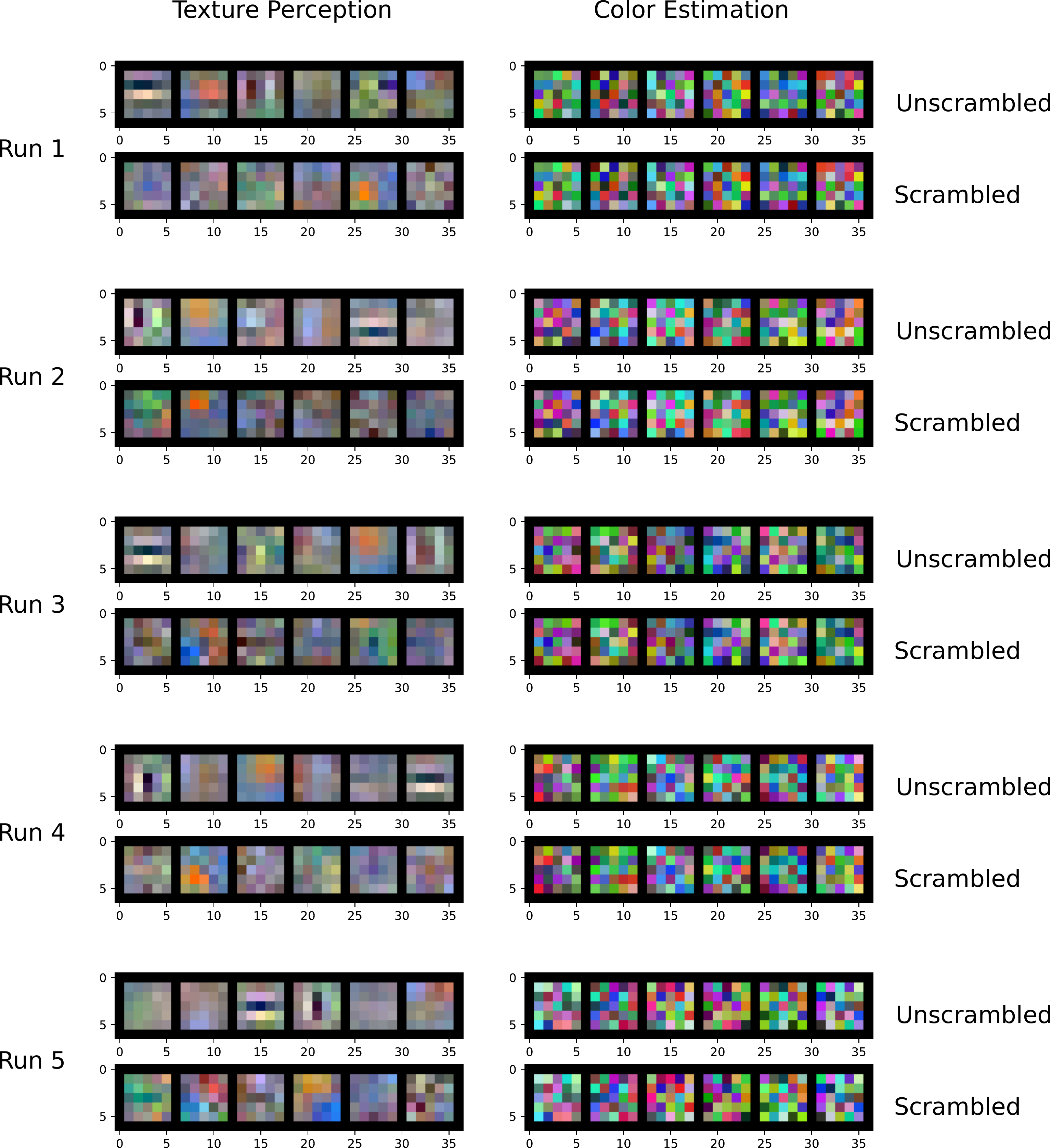}
\caption{ThreeConvNet filters for Texture Perception and Color Estimation.}
\label{fig:ThreeConvNet_B}
\end{figure*}

\textbf{Texture Perception}: Figure~\ref{fig:ThreeConvNet_B} shows that the most noticeable pattern for the learned features in the unscrambled condition is the horizontal and vertical learned patterns. In this regard, it is quite interesting to see an orthogonal filter basis emerge -- resembling the results of Hénaff~\&~Simoncelli~\cite{henaff2014local}, except that we do not add a nuclear norm-based regularizer, and this structure emerges purely driven by the texture statistics themselves. Although, this learned structure goes away for the fully scrambled textures, it is quite puzzling to see in Figure~\ref{fig:Scrambling_Texture} that the drop in generalization performance is minimal, suggesting that perhaps more complex integration structures emerge in the higher layers of the ThreeConvNet.

\textbf{Color Estimation}: Figure~\ref{fig:ThreeConvNet_B} shows no clear observable pattern other than a random distribution for both the unscrambled and scrambled images for the first layer of learned filters. 
Recall, one could be tempted to say ``no learning has occurred, which is why the filters have stayed the same'', this is incorrect as we view the convergence of the loss function for ThreeConvNet in the Color Estimation task in Figure~\ref{fig:Color_Estimation_Loss}. Interestingly, given that \textit{the filters stay the same across these two conditions} and the initial set of random weights in training is matched -- this would suggest that learning is \textit{not} tailored to the first layers of representation for the color estimation task, but rather automatically distribution to the upper (or last layers) of such networks. 

\subsection{VGG11}

\begin{figure*}[!h]
\centering
\includegraphics[width=0.5\linewidth,clip]{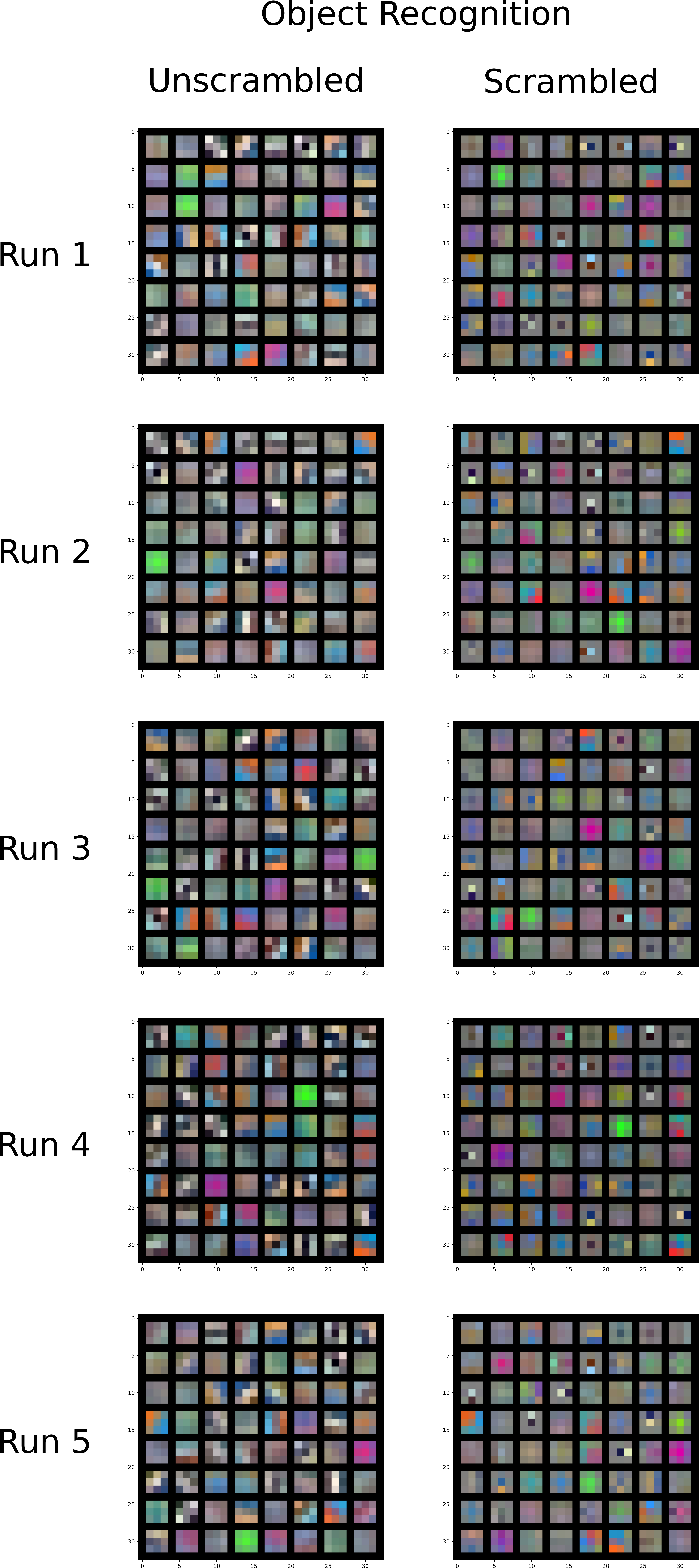}
\caption{VGG11 filters for Object Recognition}
\label{fig:VGG11_A}
\end{figure*}

\textbf{Object Recognition}: While barely noticeable given the small $3\times3$ receptive field size, we observe small edge detectors emerge at different orientations: $\{0\deg,45\deg,90\deg,135\deg\}$, both with and without color opponency across all 5 runs (Figure~\ref{fig:VGG11_A}). For the scrambled condition however, we mostly see center-surround-like operators. While we do not perform any experiments manipulating the width of VGG11 (analogous to the size of a retinal bottle-neck), this dissociation in learned representations has a resemblance to those of Lindsey~\textit{et~al.}~\cite{lindsey2019unified} (where smaller retinal bottlenecks force center-surround like computations).

\begin{figure*}[!h]
\centering
\includegraphics[width=0.5\linewidth,clip]{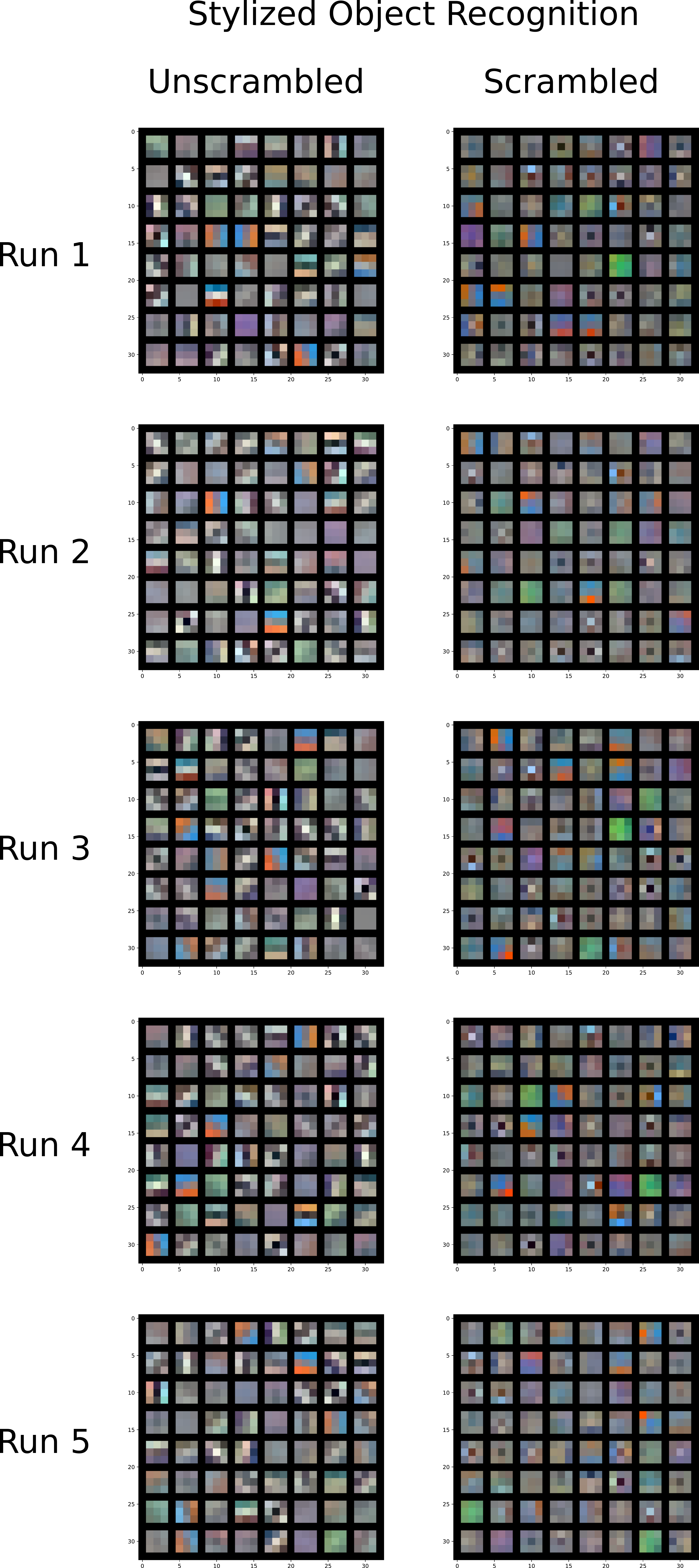}
\caption{VGG11 filters for Stylized Object Recognition}
\label{fig:VGG11_B}
\end{figure*}

\textbf{Stylized Object Recognition}:

The previous pattern of results is interestingly heavily accentuated (vs attenuated for ThreeConvNet) when we remove the texture-bias through training on the stylized CIFAR-10 dataset (Figure~\ref{fig:VGG11_B}). The number of orientation-tuned filters for the unscrambled condition and center-surround filters for the scrambled condition is even more salient, however proper quantification to justify these claims is still needed. 

\begin{figure*}[!h]
\centering
\includegraphics[width=0.5\linewidth,clip]{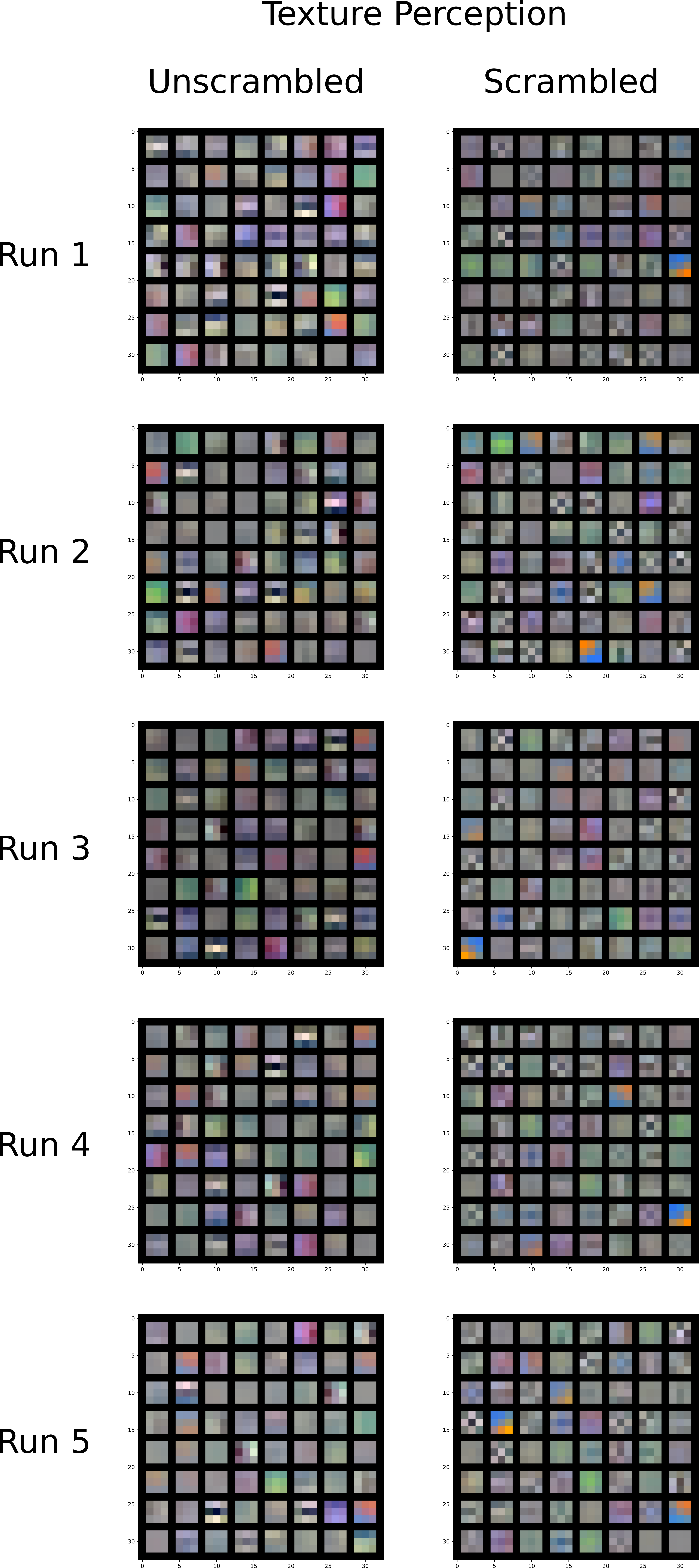}
\caption{VGG11 filters for Texture Perception}
\label{fig:VGG11_C}
\end{figure*}

\textbf{Texture Perception}: Analogous to the frequency of edge detectors learned in the object and stylized object recognition task, the texture perception task for the VGG11 has heavily relied on horizontal and vertically-tuned features (no diagonally-tuned filters) for the unscrambled condition (See Figure~\ref{fig:VGG11_C}). Similarly, the scrambled condition shows a pattern a center-surround like operators with some horizontal/vertical learned features. 

It is worth recalling that while the learned features are different, there is minimal differences in performance for the ThreeConvNet + VGG11 under both the unscrambled and scrambled condition (Figure~\ref{fig:Scrambling_Texture}).

\begin{figure*}[!h]
\centering
\includegraphics[width=0.5\linewidth,clip]{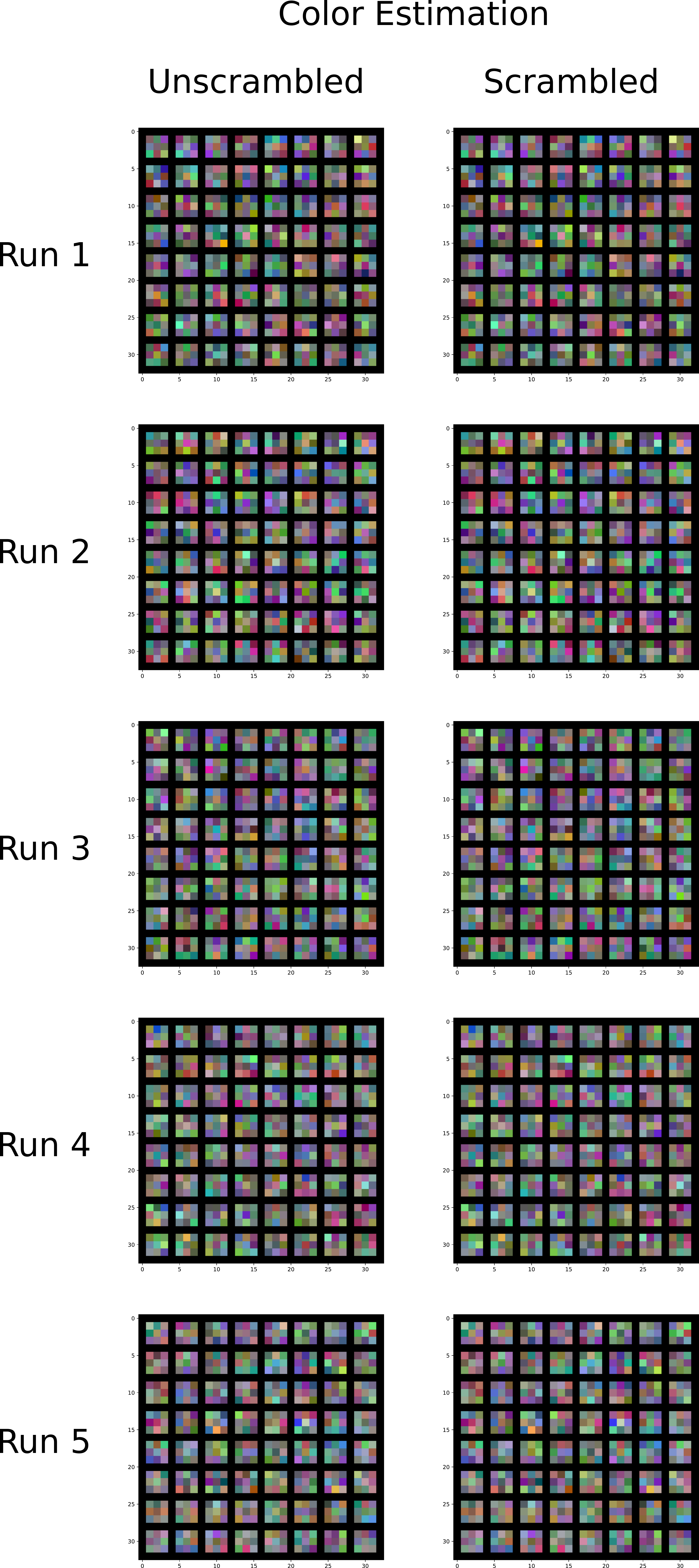}
\caption{VGG11 filters for Color Estimation}
\label{fig:VGG11_D}
\end{figure*}

\textbf{Color Estimation}: We see a similar pattern for VGG11 in Figure~\ref{fig:VGG11_D} as for ThreeConvNet, where there is a random pattern and no clear difference between the learned filters of the unscrambled and scrambled images. The same arguments apply when viewing Figure~\ref{fig:Color_Estimation_Loss}, as the VGG11 does indeed learn to estimate color, but learning is likely occurring in the later layers of processing in the network. 

It seems like the general approach for convolutional networks (ThreeConvNet + VGG11) when performing the color estimation task is to compute ``an average of the average of the average (...)'' of small receptive fields, which arrives to a close but non-optimal solution of the true color mean.

\newpage
\cleardoublepage

\newpage
\cleardoublepage

\newpage
\cleardoublepage

\newpage
\cleardoublepage

\newpage
\cleardoublepage

\newpage
\cleardoublepage
\cleardoublepage

\section{Does SGD preserve locality?}
\label{sec:Pilot}

\begin{figure}[!h]
\centering
\includegraphics[width=1.0\linewidth,clip]{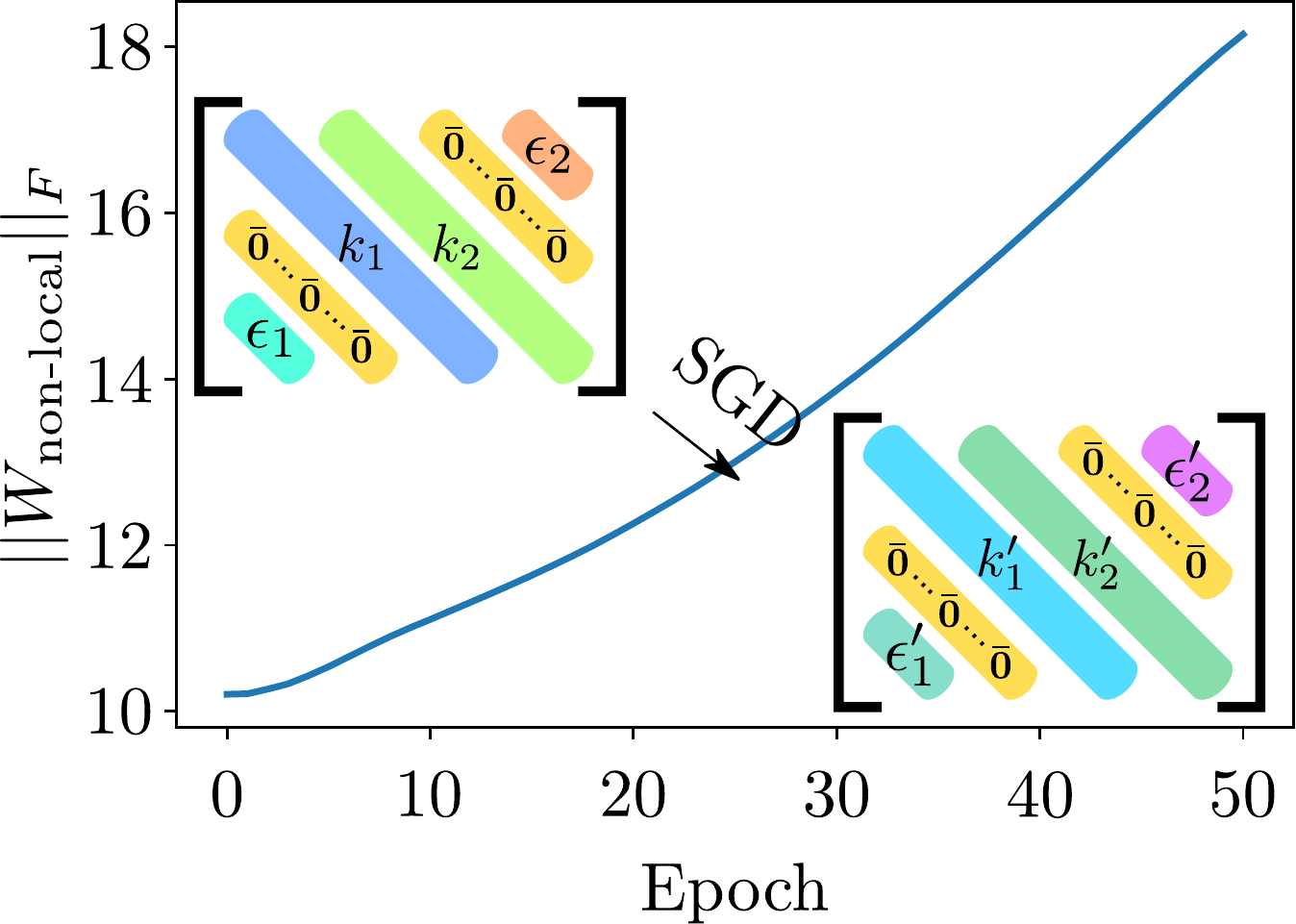}
\caption{SGD preserves non-locality.}
\label{fig:nonlocal}
\end{figure}

We performed a preliminary experiment (Figure~\ref{fig:nonlocal}) in which the architecture of a fully connected network was initialized to be a ``noisy'' convolutional one with additional nonconvolutional connections (i.e. a Toeplitz matrix with some additional non-zero entries). In Figure \ref{fig:nonlocal} we added non-zero entries to the Toeplitz matrix with probability $0.05\%$ and plotted the Frobenius norm of these additional weights during training.  We find that SGD preserves these additional weights and hence dense layers do not converge to convolutions naturally, even if initialized close to them. 

We implemented the addition of nonlocal connection to a convolutional network by constructing two networks -- one CNN and one fully connected with compatible architecture. The CNN had 4 convolutional layers with kernel size 3, stride of 1 and number of output channels 3, 6, 12, 12 respectively, followed by two fully connected layers (with 1024 hidden units) and no batch normalization layers. The models were trained with SGD with learning rate $\eta=0.01$, momentum of 0.9 and batch size of 64. 

After initialization of the CNN, we transformed the convolution layers into Toeplitz matrices and used these as initialization for the fully connected network. We then randomly added additional nonzero weights to the sparse Toeplitz matrices to implement the nonlocal connections. During training, we masked the zero entries to keep them from being updated by backpropagation steps. We added connections with probability of $0.05\%$, which resulted in increase of nonzero weights to 77005 up from 72900 for the first layer (note that in a typically initialized dense layer this would be $\sim 8.3$M parameters). In Figure \ref{fig:nonlocal} we plotted the Frobenius norm of \textit{only} the additional nonlocal connections in the first layer. We observed that this norm kept increasing throughout training, signalling that the nonlocal connections were not pruned out by SGD, agreeing with the hypothesis that fully connected layers do not converge to convolutions with standard SGD -- a result also shown recently in Neyshabur~\cite{neyshabur2020towards}.

\end{appendix}

\end{document}